\documentclass[10pt,twocolumn,letterpaper]{article}

\PassOptionsToPackage{table}{xcolor}
\usepackage{cvpr}
\usepackage{graphicx}
\usepackage{amsmath,amssymb}
\usepackage{booktabs}
\usepackage{multirow}
\usepackage{calc}
\usepackage[ruled,linesnumbered]{algorithm2e}
\usepackage[accsupp]{axessibility}
\usepackage[breaklinks,colorlinks,allcolors=blue]{hyperref}

\title{DNF-SR: Dual-Input and Negative-Aware Feature Fine-Tuning for Real-World Image Super-Resolution}

\author{Shuhao Han\textsuperscript{1,2} \quad Wenjie Liao\textsuperscript{1,2} \quad Hayden Vance\textsuperscript{3} \quad Hang Dong\textsuperscript{3} \\ Rui Zhang\textsuperscript{3} \quad Chun-Le Guo\textsuperscript{1,2} \quad Chongyi Li\textsuperscript{1,2}\thanks{Corresponding author} \\
\textsuperscript{1}VCIP, CS, Nankai University\\
\textsuperscript{2}NKIARI, Shenzhen Futian\\
\textsuperscript{3}Independent Researcher
}

\begin{document}
\maketitle
\begin{abstract}
Benefiting from the powerful generative priors of diffusion models, diffusion-based real-world image super-resolution (Real-ISR) methods have demonstrated impressive performance.
To achieve efficient Real-ISR, several recent works have designed one-step diffusion-based models.
However, unmediatedly feeding LR into a diffusion model creates a distributional gap with the model's original input.
A straightforward approach to reduce the distribution gap is to introduce noise to the LR latents. However, directly adding noise inevitably corrupts the content of the LR images.
In this study, we propose \textbf{DNF‑SR}, a \textbf{D}ual‑input and \textbf{N}egative‑aware \textbf{F}eature fine‑tuning method for Real-ISR.
Specifically, we use a \textbf{dual-input} strategy that concatenates the original LR image with the noisy LR input and feeds them into a diffusion-based image editing model, ensuring both high-fidelity one-step super-resolution and improved perceptual and content consistency.
Additionally, the noise present in the noisy LR input introduces randomness and diversity into the outputs. We exploit this property and propose a post-training optimization method, \textbf{Negative-aware Feature Fine-Tuning} (NF²T), which guides the model toward producing higher-quality results.
NF$^2$T classifies multiple outputs into positive and negative subsets and then defines implicit policy improvement directions in both the image and feature spaces, thereby further enhancing the stability of the optimization.
Extensive experiments show that DNF-SR outperforms other methods.
Code is available at \url{https://github.com/SHH-Han/DNF-SR}.
\end{abstract}

\section{Introduction}
\label{sec:intro}

\begin{figure}[t]
    \centering
    \includegraphics[width=\linewidth]{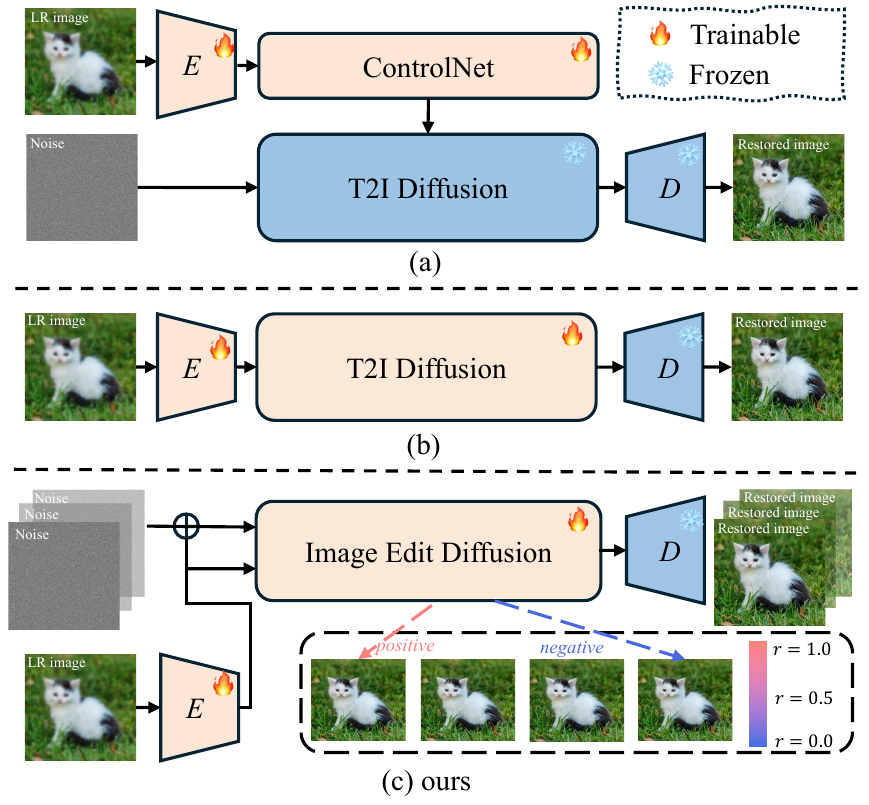}
    \caption{
Different model architectures for one‑step SR. \textbf{(a)} Utilizing ControlNet~\cite{zhang2023adding} to inject LR into a text‑to‑image (T2I) model, enabling one‑step SR from noise directly to HR.\textbf{ (b)} Directly using LR as input and fine‑tuning a T2I model to generate HR. \textbf{(c)} We adopt a dual-input method, feeding noisy LR and original LR into the image editing model to generate HR, and utilize a Negative-aware Feature Fine-Tuning method for post-training.}
    \label{fig:compare_framework}
\end{figure}
Image Super-Resolution (ISR) aims to reconstruct a clear high-resolution (HR) image from a degraded low-resolution (LR) image suffering from noise, blur, and other degradations.
To better reconstruct low-quality real-world images that are affected by a wider range of degradations into more realistic images, many researchers have begun exploring the use of generative models \cite{chen2023pixart, saharia2022photorealistic} for image super-resolution.
Recently, as large-scale pretrained diffusion generative models \cite{rombach2022high, podell2023sdxl} have demonstrated impressive performance in image generation, an increasing number of studies \cite{sun2023improving, lin2024diffbir, wu2024seesr, duan2025dit4sr} have adopted techniques such as LoRA \cite{hu2022lora} or ControlNet \cite{zhang2023adding} to transfer the powerful high-quality image generation priors of diffusion models to Real-ISR, yielding more realistic and clearer restored images.
Meanwhile, optimization under data constraints has also been explored \cite{cao2025analytical}.

Due to the multi-step denoising process inherent to diffusion models, they suffer from slow inference.
To better leverage diffusion models for efficient super-resolution, recent works \cite{wang2024sinsr, wu2024osediff, zhang2024degradation, dong2025tsd, lin2025hypir, sun2025pixel, wu2025omgsr} have explored one-step diffusion-based  SR models.
As shown in Fig. \ref{fig:compare_framework}, the current mainstream frameworks for one-step diffusion-based SR methods can be broadly categorized into the following two types.
(a) Distilling multi-step super-resolution diffusion networks that use ControlNet (e.g., resshift \cite{yue2024resshift}, seesr \cite{wu2024seesr}) into a one-step mapping from noise to high-resolution images, as in SinSR \cite{wang2024sinsr} and AddSR-1s \cite{xie2024addsr}.
These methods increase the parameter count by incorporating ControlNet.
And the design that directly restores an HR image from noise in a single step limits the model’s performance.
(b) Directly feeding the LR image into the diffusion model and fine-tuning the model with LoRA to reconstruct a clear image has been explored by works such as OSEDiff \cite{wu2024osediff}, TSDSR \cite{dong2025tsd}, and FluxSR \cite{li2025fluxsr}.
These methods replace the diffusion model’s original noise input at the initial timesteps of the denoising process with the LR image latent, which introduces a substantial distributional gap in the model inputs and consequently degrades performance.
Recent works such as OMGSR \cite{wu2025omgsr} and TADSR \cite{zhang2025timeawarestepdiffusionnetwork}, which still follow the (b) architecture, explore using intermediate timesteps to perform one-step diffusion super resolution.
However, both the HR latent and the noise contain more high-frequency information than the LR latent.
Therefore, at any intermediate timestep, the LR latent differs from the diffusion model's original input in terms of high-frequency information.
Adding noise to LR is the most straightforward method to reduce the distribution gap, but it may destroy the content of LR.

To address the issues above, we propose \textbf{DNF-SR}, a \textbf{D}ual-input and \textbf{N}egative-aware \textbf{F}eature Fine-Tuning method for Real-ISR, with its framework shown in Fig.~\ref{fig:compare_framework}(c).
Specifically, we design a \textbf{dual-input } strategy that inputs noisy LR to eliminate the gap with the original input of the Diffusion model, while additionally incorporating the original LR as a condition to ensure fidelity.
And using Flux-Kontext \cite{batifol2025flux} as a pretrained diffusion model, we patchify both the noisy LR and the original LR, then concatenate them along the token dimension and feed these tokens into the DiT~\cite{peebles2023scalable} blocks for image super-resolution.
With this dual-input design, our method reduces the input distribution gap of the diffusion model, more effectively leverages the generative prior of the diffusion model, and preserves the fidelity of the LR image.
Rather than employing the text‑to‑image generation model Flux, using an editing model can better perceive the original LR content as a condition, thereby improving the quality of outputs.

Due to the introduction of noise, our one-step SR model can generate diverse restored images.
To further improve the quality of generated images, we propose a novel super-resolution post-training method named \textbf{N}egative-aware \textbf{F}eature \textbf{F}ine-\textbf{T}uning, named \textbf{NF$^2$T}.
NF$^2$T draws on the approach of DiffusionNFT \cite{zheng2025diffusionnft} by partitioning the multiple generated images into positive and negative subsets according to a reward model and then defining an implicit policy improvement direction to enhance model performance.
However, we observed that applying DiffusionNFT to the super‑resolution task introduces pronounced grid artifacts in the reconstructed images due to the implicit optimization direction in the latent space.
Therefore, we adopt NF$^2$T, transferring the optimization from the latent space to the image and the feature spaces for improved performance.
Compared with other reinforcement learning methods \cite{wallace2024diffusion,liu2025flow}, NF$^2$T omits probabilistic modeling in sampling and optimization, improving generative model performance by directly optimizing the predicted velocity in flow matching.
Meanwhile, NF$^2$T can be optimized exploiting multiple sampled images, unlike DiffusionDPO, which only uses image pairs.
During post-training, we evaluate the multiple generated images using several IQA methods \cite{blau2018perception, ding2020image, ke2021musiq, yang2022maniqa, wang2023clipiqa}, including full reference (FR)  and no reference (NR) metrics, and then aggregate the normalized scores into a single reward for optimization.
The use of diverse metrics provides clearer optimization signals for the model and thereby enables recovery of higher quality images.

The main contributions of our work are as follows:
\begin{itemize}
    \item We design a dual-input strategy that feeds both noisy LR and original LR into an image‑editing model, which can effectively capture conditional LR image information. This approach better leverages the model's generative prior while ensuring higher fidelity, thereby further enhancing the performance of SR model.
    \item We propose a novel SR optimization method named Negative-aware Feature Fine-Tuning. It categorizes multiple sampled restored images into positive and negative optimization directions within the image and feature spaces, which further improves the quality of outputs.
\end{itemize}

\section{Related Work}
\label{sec:related}

\noindent \textbf{Real-World Image Super-Resolution.}
Recently, large-scale pretrained diffusion generative models \cite{rombach2022high, podell2023sdxl} have achieved strong results in Real-ISR tasks \cite{sun2023improving, lin2024diffbir, wu2024seesr, duan2025dit4sr, wu2025exploring, wang2024super, yang2025jvcsr+, song2025wdfsr}, yielding more realistic and clearer restored images.
However, the multi-step denoising process results in high computational and time costs. Consequently, one-step Real-ISR models have become a primary research focus for faster inference.
OSEDiff \cite{wu2024osediff} introduces the VSD loss to distill the pre-trained SD model. Building upon this, TSD-SR \cite{dong2025tsd} designs a Target Score Distillation for SR.
SinSR \cite{wang2024sinsr} achieves one-step ISR from noise to HR by distilling a multi-step diffusion model into a student network through deterministic mapping.
OMGSR \cite{wu2025omgsr} injects the LR image latent distribution at a pre-computed mid-timestep, achieving the state-of-the-art performance.
Nevertheless, there is a gap between the LR latent and the diffusion model’s original input in terms of high-frequency information.
To eliminate this gap, we design dual-input methods, inputting noisy LR and incorporating original LR as a condition.  Furthermore, an image-editing model is employed to leverage this dual-path input better, thereby producing higher‑quality outputs.

\noindent \textbf{Preference Alignment for Diffusion Models.}
Since the proposal of GRPO \cite{guo2025deepseek}, Reinforcement Learning from Human Feedback (RLHF) \cite{ouyang2022training, bai2022training}  has emerged as one of the most prominent and widely studied research topics.
Diffusion models and rectified flows can also benefit significantly from alignment with human feedback for their generation diversity.
Many methods are based on likelihood estimation, including:
(1) Policy gradient methods from PPO-style \cite{black2023training, fan2023dpok} decompose trajectory likelihoods stepwise without forward consistency, while recent GRPO extensions \cite{liu2025flow, xue2025dancegrpo} convert ODE to SDE samplers, proving effective and scalable for diffusion RL.
However, this method is hard to apply to one-step Real-ISR diffusion since it requires a multi-step denoising process to guarantee diversity.
(2) Direct Policy Optimization (DPO)-style \cite{rafailov2023direct} methods. Diffusion-DPO \cite{wallace2024diffusion} adapts DPO to diffusion, and it is inherently confined to paired human preference data, whose annotation process is labor-intensive and expensive.
Moreover, its pairwise training paradigm hinders the exploitation of accurate ranking information among multiple generated results, thereby limiting the speed of optimization.
Recently, DiffusionNFT \cite{zheng2025diffusionnft} eliminates the reliance on likelihood estimation and SDE-based reverse process by formulating policy improvement as a contrast between positive and negative generations.
It fully utilizes information from multiple sampled results to determine a better optimization direction.
However, directly applying DiffusionNFT to SR tasks introduces a pronounced grid artifact.
To address this issue, we propose a novel Negative-aware feature fine-tune (NF$^2$T) method.

\section{Methodology}

\begin{figure*}[t]
    \centering
    \includegraphics[width=\linewidth]{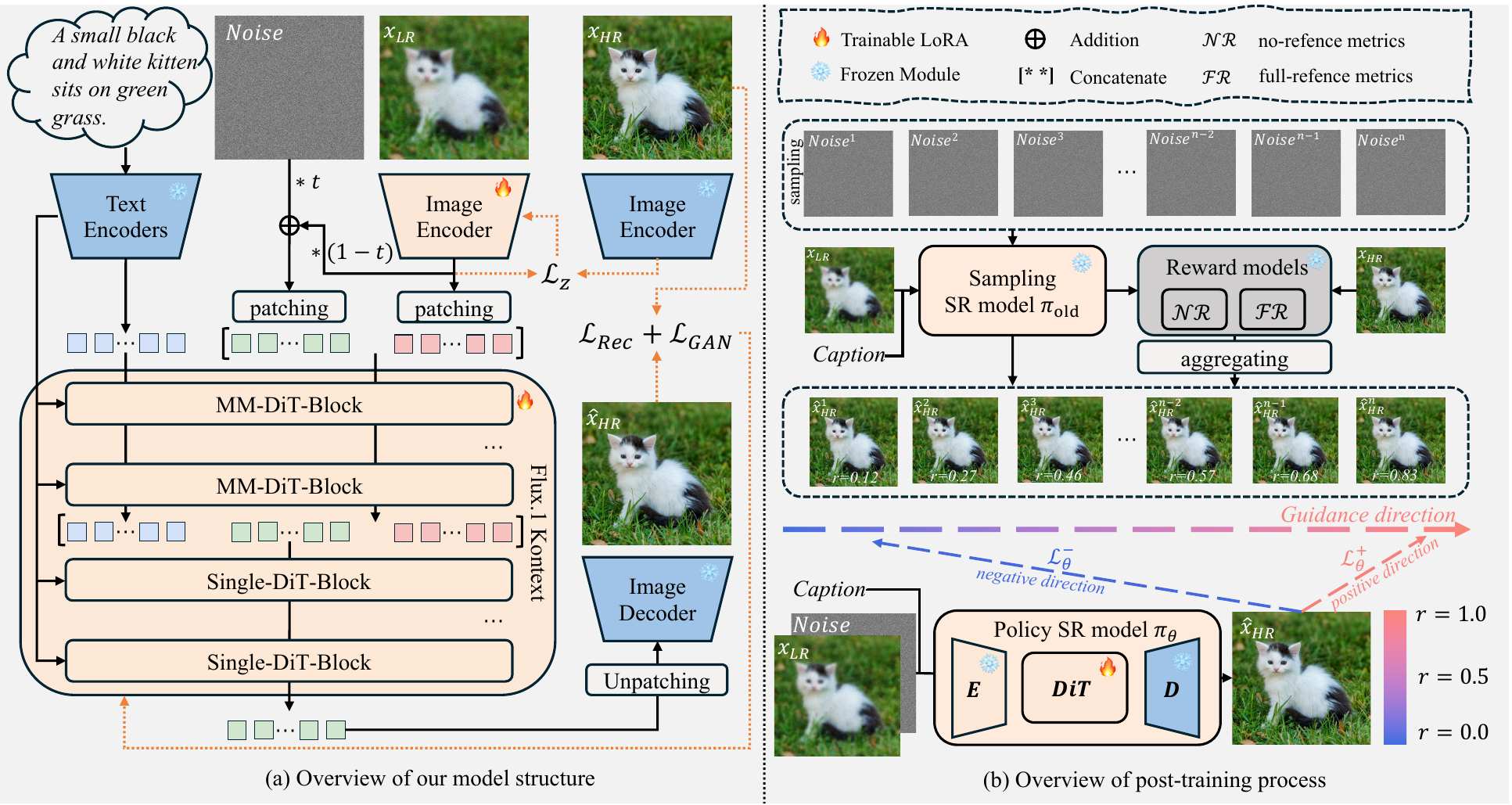}
    \caption{ Overview of the DNF-SR framework. (a) The model structure adopts a dual-input design: noisy LR latents (\(z_{mix}\)) and original LR latents (\(z_{LR}\)) are concatenated with text tokens (encoded from image captions) and fed into fine-tuned DiT blocks of the Flux-Kontext pretrained model. We employed multiple loss functions to ensure the performance of the model. (b)The post-training process of Negative-aware Feature Fine-Tuning: multiple restored results are generated with different input noises, evaluated by combined full-reference (FR) and no-reference (NR) IQA metrics to obtain reward scores, and divided into positive and negative subsets. By constructing positive $\mathcal{L}_\theta^+$ and negative $\mathcal{L}_\theta^-$ optimization directions within both the image and the feature spaces, the quality of the restored images is improved.}
    \label{fig:framework}
\end{figure*}

\subsection{Preliminaries}
\textbf{Flow-matching Model in Flux.}
As a core generative model used by Flux \cite{blackforestlabs2024}, flow matching (FM) \cite{lipman2023flow} aims to learn a continuous normalizing flow that maps a simple distribution (e.g., standard Gaussian distribution $\epsilon \in \mathcal{N}(0, \mathbf{I}) $) to the target data distribution $x_{0} \sim \pi_{0}=p_{data}$.
The specific training loss function is as follows:
\begin{equation}
    \mathcal{L}=\mathbb{E}_{x_{0} \sim \pi_{0}, \epsilon \sim \mathcal{N}(0,\mathbf{I}), t}||v_{\theta}(x_t,t)-v||^2,
    \label{eq:fm_loss}
\end{equation}
where $v$ denotes the velocity at any time $t \in [0,1]$ predicted by the model. Flux adopts the same method from Rectified Flow \cite{liuflow}, which assumes a simple linear path $x_t=(1-t)x_0+t\epsilon$, thereby deriving the optimization target for velocity as $v=\epsilon-x_0$.

\noindent \textbf{DiffusionNFT for Preference Alignment.}
DiffusionNFT adopts a novel Preference Alignment Paradigm for diffusion models to achieve efficient alignment with human or task-specific preferences.
The specific approach of DiffusionNFT involves computing the reward $r$ for multiple sampled data from the diffusion model, normalizing these rewards to the range $[0,1]$, then leveraging positive and negative sample signals to implicitly model preference directions.
The specific training objective is as follows:
\begin{equation}
\begin{split}
    \mathcal{L}=\mathbb{E}_{x_{0} \sim \pi_{old}(x_{0}|c),\epsilon \sim \mathcal{N}(0,\mathbf{I}),c, t}  \Big[ r||v_{\theta}^+(x_t,c,t) - v||^2 \\  + (1-r) ||v_{\theta}^{-}(x_t,c,t)-v||^2 \Big],
\end{split}
\label{eq:NFT}
\end{equation}
where implicit positive policy:
\begin{align*}
{v}_\theta^+({x}_t, {c}, t) := (1 - \beta){v}^{\text{old}}({x}_t, {c}, t) + \beta{v}_\theta({x}_t, {c}, t), \quad
\end{align*}
and implicit negative policy:
\begin{align*}
    {v}_\theta^-({x}_t, {c}, t) := (1 + \beta){v}^{\text{old}}({x}_t, {c}, t) - \beta{v}_\theta({x}_t, {c}, t).
\end{align*}
In Eq. \eqref{eq:NFT}, $\pi_{\text{old}}$ denotes the pretrained diffusion policy, $v_{\text{old}}$ and $v_\theta$ represent the velocity predictor of the pretrained diffusion and the optimized diffusion model, respectively, $c$ denotes conditional control, and $\beta$ is a hyperparameter representing the guidance strength.

\subsection{Overview of DNF-SR}
As shown in Fig. \ref{fig:framework}, DNF-SR mainly consists of two parts. One part is the \textbf{model structure of DNF-SR}, where we use Flux-kontext as a pre-trained model and design a single-step super-resolution model with a dual-input at a mid-timestep $t_{mid}$. The other part is the \textbf{post-training of DNF-SR}, where we adopt a Negatively-aware Feature Fine-Tuning (\textbf{NF$^2$T}) method, achieving preference optimization by dividing sampled data into positive and negative optimization direction in both image and feature spaces.

\subsection{Model Structure of DNF-SR}
The model structure of DNF-SR is shown in Fig. \ref{fig:framework}(a).
The inputs to DNF-SR are the LR image $x_{LR}$, its caption $c$, and randomly initialized noise $\epsilon$.
First, $x_{LR}$ is fed into a fine-tuned Encoder $E_{\theta}$ to obtaion $z_{LR}$.
Then $z_{LR}$ and $\epsilon$ are weighted according to a fixed mid-timestep $t_{mid}$ to get $z_{mix} = (1-t_{mid}) z_{LR} + t_{mid} \epsilon$.
Next, DNF-SR feeds the concatenated $z_{mix}$ and $z_{LR}$ along with the text condition $z_{text}$ (encoded using c) into fine-tuned DiT blocks, as employed in Flux-Kontext.
Specifically,  all tokens from $z_{mix}$, $z_{LR}$, and $z_{text}$ are concatenated after passing through multiple MM-DiT blocks. This combined sequence then goes through several Single-DiT blocks.
Subsequently, the output at the position corresponding to $z_{mix}$ is extracted as the model's predicted velocity $v_{\theta}$, which is then used to obtain $\hat{z}_{HR}=z_{mix}-t_{mid}v_{\theta}$.
Finally, $\hat{z}_{HR}$ is passed through a fixed-parameter decoder $D_{\varphi}$ to obtain $\hat{x}_{HR}$.

In comparison to other methods, our primary improvements involve employing a \textbf{dual input for SR} and performing a \textbf{one-step denoising at mid-timestep}.

\noindent \textbf{Dual-input for SR.}
Most current single-step SR methods directly feed the LR image $x_{LR}$ into a Diffusion model to obtain the HR image $\hat{x}_{HR}$.
Compared to the original noise input of Diffusion models, there's a significant distribution gap between the LR data and noise.
Directly replacing the input makes it difficult for Diffusion's pre-trained parameters to be effectively applied to SR tasks.
Although recent works \cite{wu2025omgsr, zhang2025timeawarestepdiffusionnetwork} have explored replacing the input at mid-timestep to mitigate this gap between LR and the Diffusion model's original input, this introduces a new question: \textit{At which timestep $t$ does the LR latent $z_{LR}$ have a smaller distribution difference with the original Diffusion model's latent input $z_t$? }
In fact, from a frequency perspective, $z_{LR}$ is most consistent with the latent representation of a natural image $z_0$ at $t=0$, rather than an arbitrary mid-timestep's $z_t$.
Relative to \(x_0\), \(x_t = (1-t)x_0 + t\epsilon\) exhibits richer high-frequency components owing to noise injection. For super-resolution, low-resolution (LR) images contain fewer high-frequency components than high-resolution (HR) images \(x_0\). Thus, \(x_0\) is distributionally more similar to LR images in the diffusion framework.
However, in SR tasks,  if $t=0$, the model would be unable to optimize effectively due to $\hat{z}_{HR} = z_{LR} - tv_{\theta}$.
To address this issue and better eliminate the distribution gap between $z_{LR}$ and $z_t$, one straightforward method is adding noise to $z_{LR}$ to obtain $z_{mix}$ as the input to the model. However, it can corrupt the content of the LR images.
Therefore, to avoid content degradation caused by adding noise to $z_{LR}$, we design a dual-input method, which feeds the original $z_{LR}$ as a conditional control along with $z_{mix}$ into the Diffusion model.
Meanwhile, unlike other single‑step super‑resolution approaches, which use text‑to‑image models as the pretrained backbone, we employ a pretrained image‑editing model. This enables more effective utilization of the original LR image as a conditioning signal, thereby further enhancing super‑resolution performance.
Through this approach, we better leverage the model's pre-trained parameters while maintaining image fidelity, generating high-quality $\hat{x}_{HR}$.

\noindent \textbf{One-step denoising at mid-timestep.}
Diffusion models focus on generating different frequency components at various timesteps. Specifically, as $t$ decreases, the model's attention shifts from generating low-frequency information (i.e., the overall structure of the image) towards prioritizing the generation of high-frequency information (i.e., fine-grained details and textures).
To ensure both fidelity and visual quality, we choose the latent $z_{mix}$ corresponding to an intermediate timestep $t_{mid}$ as the model input, rather than pure noise $\epsilon$.

\begin{figure*}
    \centering
    \includegraphics[width=\textwidth]{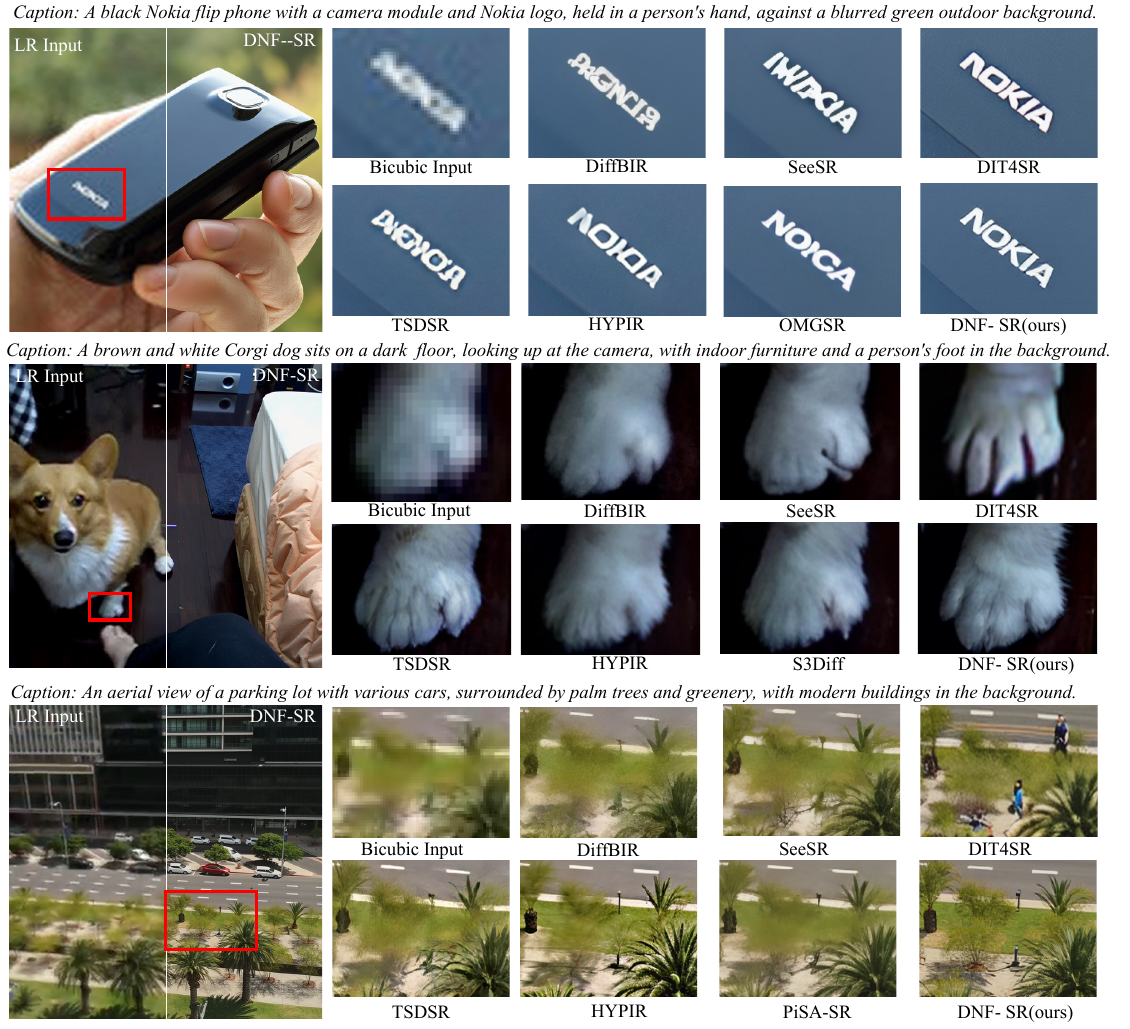}
    \caption{Visual comparisons of different Real-ISR methods. Please zoom in for a better view.}
    \label{fig:visual_all}
\end{figure*}

\begin{table*}[htbp]
  \centering
  \caption{Quantitative comparison of DNF-SR with SOTA Real-ISR methods on four datasets. DNF-SR(sft) representing DNF-SR only undergoing supervised fine-tuning (without \(NF^2T\) post-training) and DNF-SR denoting the full model with \(NF^2T\).}
  \label{tab:main_tab}
  \resizebox{\textwidth}{!}{
    \begin{tabular}{lc|ccc|ccccccccc}
    \toprule
    \multicolumn{1}{l}{\multirow{2}{*}{\textbf{Datasets}}} & \multirow{2}[3]{*}{\textbf{Metrics}} & \multicolumn{3}{c|}{\textbf{Multi-step Methods}} & \multicolumn{9}{c}{\textbf{One-step Methods}} \\
    \cmidrule{3-14}
    \multicolumn{1}{l}{} &       & \textbf{DiffBIR} & \textbf{SeeSR} & \textbf{DiT4SR} & \textbf{SinSR-1s} & \textbf{OSEDiff} & \textbf{S3Diff} & \textbf{PisaSR} & \textbf{TSDSR} & \textbf{HYPIR} & \textbf{OMGSR} & \textbf{DNF-SR(sft)} & \textbf{DNF-SR} \\
    \midrule
    \multirow{7}{*}{\textbf{RealSR}}
    & PSNR↑  & 24.835  & 25.147  & 23.479  & \textcolor[rgb]{ 1,  0,  0}{\textbf{26.166}}  & 25.141  & 25.183  & 25.503  & 23.404  & 22.785  & 25.882  & 25.628  & 24.970  \\
    & LPIPS↓ & 0.3650  & 0.3007  & 0.3154  & 0.3062  & 0.3209  & \underline{\textcolor[rgb]{ 0,  0,  1}{0.2721}}  & \textcolor[rgb]{ 1,  0,  0}{\textbf{0.2672}}  & 0.2805  & 0.3107  & 0.2779  & 0.2925  & 0.3239  \\
    & CLIPIQA↑ & 0.7054  & 0.6706  & 0.6319  & 0.6243  & 0.6613  & 0.6732  & 0.6698  & \underline{\textcolor[rgb]{ 0,  0,  1}{0.7196}}  & 0.6491  & 0.6682  & 0.6903  & \textcolor[rgb]{ 1,  0,  0}{\textbf{0.7257}}  \\
    & MUSIQ↑ & 69.279  & 69.822  & 68.197  & 61.370  & 67.263  & 67.828  & 70.149  & \underline{\textcolor[rgb]{ 0,  0,  1}{70.766}}  & 66.559  & 69.527  & 70.672  & \textcolor[rgb]{ 1,  0,  0}{\textbf{72.040}}  \\
    & MANIQA↑ & 0.6502  & 0.6451  & 0.6594  & 0.5418  & 0.6317  & 0.6424  & 0.6551  & 0.6312  & 0.6558  & 0.6695  & 0.6856  & \textcolor[rgb]{ 1,  0,  0}{\textbf{0.6930}}  \\
    & QALIGN↑ & 3.7894  & 3.7187  & 3.3974  & 3.1774  & 3.6634  & 3.6632  & 3.6334  & 3.7754  & 3.6931  & 3.8507  & \underline{\textcolor[rgb]{ 0,  0,  1}{3.9162}}  & \textcolor[rgb]{ 1,  0,  0}{\textbf{4.0718}}  \\
    & VQ-R1↑ & 3.9388  & 3.7811  & 3.7111  & 3.2470  & 3.9522  & 3.9541  & 3.8162  & 3.8224  & 3.9600  & 4.1285  & \underline{\textcolor[rgb]{ 0,  0,  1}{4.1310}}  & \textcolor[rgb]{ 1,  0,  0}{\textbf{4.2646}}  \\
    \midrule
    \multirow{7}{*}{\textbf{DrealSR}}
    & PSNR↑  & 25.904  & 28.070  & 25.681  & 28.149  & 27.924  & 27.539  & \underline{\textcolor[rgb]{ 0,  0,  1}{28.319}}  & 26.197  & 25.899  & \textcolor[rgb]{ 1,  0,  0}{\textbf{28.928}}  & 28.254  & 28.141  \\
    & LPIPS↓ & 0.4670  & 0.3174  & 0.3693  & 0.3479  & \underline{\textcolor[rgb]{ 0,  0,  1}{0.2966}}  & 0.3109  & 0.2960  & 0.3115  & 0.3391  & \textcolor[rgb]{ 1,  0,  0}{\textbf{0.2952}}  & 0.3210  & 0.3531  \\
    & CLIPIQA↑ & 0.7065  & 0.6910  & 0.6658  & 0.6564  & 0.6963  & 0.7133  & 0.6971  & \underline{\textcolor[rgb]{ 0,  0,  1}{0.7302}}  & 0.6429  & 0.6914  & 0.6914  & \textcolor[rgb]{ 1,  0,  0}{\textbf{0.7559}}  \\
    & MUSIQ↑ & 66.137  & 65.085  & 64.861  & 57.242  & 64.692  & 63.955  & 66.108  & 66.120  & 61.084  & 65.862  & \underline{\textcolor[rgb]{ 0,  0,  1}{67.247}}  & \textcolor[rgb]{ 1,  0,  0}{\textbf{68.732}}  \\
    & MANIQA↑ & 0.6221  & 0.6052  & 0.6253  & 0.5027  & 0.5898  & 0.6124  & 0.6160  & 0.5820  & 0.6055  & 0.6315  & \textcolor[rgb]{ 1,  0,  0}{\textbf{0.6542}}  & \underline{\textcolor[rgb]{ 0,  0,  1}{0.6515}}  \\
    & QALIGN↑ & 3.7142  & 3.5867  & 3.3592  & 3.1929  & 3.5434  & 3.6155  & 3.5828  & 3.6943  & 3.5447  & 3.6877  & \underline{\textcolor[rgb]{ 0,  0,  1}{3.7166}}  & \textcolor[rgb]{ 1,  0,  0}{\textbf{3.7997}}  \\
    & VQ-R1↑ & 3.6708  & 3.4905  & 3.4919  & 3.1672  & 3.6619  & 3.6538  & 3.5911  & 3.6615  & 3.6865  & \underline{\textcolor[rgb]{ 0,  0,  1}{3.8938}}  & 3.8478  & \textcolor[rgb]{ 1,  0,  0}{\textbf{3.9152}}  \\
    \midrule
    \multirow{7}{*}{\textbf{DIV2K}}
    & PSNR↑  & 23.145  & 23.678  & 21.789  & \textcolor[rgb]{ 1,  0,  0}{\textbf{24.291}}  & 23.724  & 23.530  & 23.867  & 22.173  & 22.252  & \underline{\textcolor[rgb]{ 0,  0,  1}{24.050}}  & 23.524  & 23.631  \\
    & LPIPS↓ & 0.3669  & 0.3194  & 0.3490  & 0.3226  & 0.2942  & \textcolor[rgb]{ 1,  0,  0}{\textbf{0.2581}}  & 0.2823  & \underline{\textcolor[rgb]{ 0,  0,  1}{0.2736}}  & 0.3042  & 0.2924  & 0.3030  & 0.3234  \\
    & CLIPIQA↑ & \underline{\textcolor[rgb]{ 0,  0,  1}{0.7300}}  & 0.6936  & 0.6636  & 0.6532  & 0.6680  & 0.7001  & 0.6927  & 0.7149  & 0.6512  & 0.6828  & 0.7028  & \textcolor[rgb]{ 1,  0,  0}{\textbf{0.7723}}  \\
    & MUSIQ↑ & 69.872  & 68.672  & 68.039  & 63.276  & 67.963  & 67.924  & 69.679  & \underline{\textcolor[rgb]{ 0,  0,  1}{70.651}}  & 65.644  & 68.624  & 69.889  & \textcolor[rgb]{ 1,  0,  0}{\textbf{71.546}}  \\
    & MANIQA↑ & 0.6440  & 0.6222  & 0.6413  & 0.5410  & 0.6131  & 0.6311  & 0.6375  & 0.6077  & 0.6227  & 0.6500  & \textcolor[rgb]{ 1,  0,  0}{\textbf{0.7009}}  & \underline{\textcolor[rgb]{ 0,  0,  1}{0.6703}}  \\
    & QALIGN↑ & \underline{\textcolor[rgb]{ 0,  0,  1}{4.1015}}  & 3.9766  & 3.7256  & 3.5216  & 3.8352  & 3.8664  & 3.8806  & 3.9271  & 3.8159  & 3.9967  & 4.0604  & \textcolor[rgb]{ 1,  0,  0}{\textbf{4.1563}}  \\
    & VQ-R1↑ & 4.0356  & 4.0486  & 3.9834  & 3.3441  & 4.0326  & 3.9861  & 4.0576  & 3.8899  & 3.9462  & 4.2884  & \underline{\textcolor[rgb]{ 0,  0,  1}{4.2899}}  & \textcolor[rgb]{ 1,  0,  0}{\textbf{4.3630}}  \\
    \midrule
    \multirow{5}{*}{\textbf{RealLQ250}}
    & CLIPIQA↑ & 0.7137  & 0.7031  & 0.7338  & 0.7140  & 0.6724  & 0.7044  & 0.7055  & 0.7219  & 0.6875  & 0.7435  & \underline{\textcolor[rgb]{ 0,  0,  1}{0.7683}}  & \textcolor[rgb]{ 1,  0,  0}{\textbf{0.7997}}  \\
    & MUSIQ↑ & 67.531  & 70.860  & 72.085  & 65.305  & 69.556  & 69.193  & 71.245  & 72.099  & 68.991  & 71.975  & \underline{\textcolor[rgb]{ 0,  0,  1}{73.077}}  & \textcolor[rgb]{ 1,  0,  0}{\textbf{73.700}}  \\
    & MANIQA↑ & 0.5878  & 0.6001  & 0.6826  & 0.5258  & 0.5782  & 0.6016  & 0.6053  & 0.5829  & 0.6044  & 0.6864  & \underline{\textcolor[rgb]{ 0,  0,  1}{0.6990}}  & \textcolor[rgb]{ 1,  0,  0}{\textbf{0.7029}}  \\
    & QALIGN↑ & 3.9757  & 4.1543  & 3.9903  & 3.7532  & 4.2481  & 4.2903  & 4.2070  & 4.1687  & 4.2316  & 4.3073  & \underline{\textcolor[rgb]{ 0,  0,  1}{4.4017}}  & \textcolor[rgb]{ 1,  0,  0}{\textbf{4.4752}}  \\
    & VQ-R1↑ & 4.1460  & 4.3657  & 4.2968  & 3.6513  & 4.4707  & 4.3840  & 4.4426  & 4.2756  & 4.4300  & 4.4951  & \underline{\textcolor[rgb]{ 0,  0,  1}{4.5706}}  & \textcolor[rgb]{ 1,  0,  0}{\textbf{4.6090}}  \\
    \bottomrule
    \end{tabular}
  }
\end{table*}

\noindent \textbf{Training Objective.}
As shown in Fig.~\ref{fig:framework}, the loss functions during the fine-tuning stage of DNF-SR consist of three components: $\mathcal{L}_z$, $\mathcal{L}_{Rec}$, and $\mathcal{L}_{GAN}$.
Specifically, $\mathcal{L}_z$ is designed to align the LR latent $z_{LR}$, after passing through the fine-tuned Encoder $E_{\theta}$, with the origin HR latent $z_{HR}$. This is formulated as:
\begin{equation}
    \mathcal{L}_z = \mathcal{L}_{MSE}(z_{LR}, z_{HR}).
\end{equation}
$\mathcal{L}_{Rec}$ represents the reconstruction loss between the restored $\hat{x}_{HR}$ and the ground-truth HR image $x_{HR}$, calculated as:
\begin{equation}
\begin{split}
    \mathcal{L}_{Rec}(\hat{x}_{HR},x_{HR}) = \mathcal{L}_{MSE}(\hat{x}_{HR},x_{HR})\\ + \mathcal{L}_{DISTS} (\hat{x}_{HR},x_{HR}).
\end{split}
\end{equation}
$\mathcal{L}_{GAN}$ is the generative adversarial~\cite{goodfellow2020generative}, where we employ DINOv3 as the discriminator.
Ultimately, the total loss is:
\begin{equation}
    \mathcal{L}_{sft} = \lambda_1\mathcal{L}_{z} + \lambda_2\mathcal{L}_{MSE} + \lambda_3\mathcal{L}_{DISTS} + \lambda_4\mathcal{L}_{GAN}.
    \label{sft_loss}
\end{equation}

\subsection{Post-training of DNF-SR}
To further enhance the quality of the reconstructed images, we adopt DiffusionNFT, a preference optimization method based on Negative-aware Fine-Tuning (NFT), to improve model performance.

When applying DiffusionNFT, according to Eq.~\eqref{eq:NFT}, we set the guidance strength $\beta=1$ and use a fixed mid-timestep $t_{mid}$, resulting in the NFT optimization objective for single-step super-resolution as:
\begin{equation}
\begin{split}
    \mathcal{L}_{NFT}=\mathbb{E}_{\pi_{old}(\hat{z}_{HR}|z_{LR}),z_{LR}}  \Big[  r||v_{\theta}^+(z_{mix},z_{LR}) - v||^2 \\ + (1-r) ||v_{\theta}^{-}(z_{mix},z_{LR})-v||^2 \Big],
\end{split}
\label{eq:NFT_sr}
\end{equation}
where $z_{mix}=t_{mid}\epsilon + (1-t_{mid})z_{LR}$, $v=(z_{mix}-\hat{z}_{HR})/t_{mid}$, $v_{\theta}^+=v_{\theta}$, $v_{\theta}^-=2v_{old}-v_{\theta}$ and $r$ represents the reward of $\hat{x}_{HR}$.
Simultaneously, we define $\mathcal{L}_{\theta}^+$ to represent the positive optimization objective
and $\mathcal{L}_{\theta}^-$ to represent the negative optimization objective:
\begin{equation}
    \mathcal{L}_{\theta}^+=r||v_{\theta}^+-v||^2,\mathcal{L}_{\theta}^-=(1-r)||v_{\theta}^--v||^2.
\end{equation}

\noindent \textbf{Feature Space Optimization.} We observe that directly using $\mathcal{L}_{\theta}^+$ and $\mathcal{L}_{\theta}^-$ as optimization objective led to noticeable grid artifacts in the output images of the optimized model. Therefore, we transform the optimization in the velocity space to the image space using a function $f(v)=D_{\varphi}(z_{mix}-t_{mid}v)$. $v_\theta^+,v_\theta^-,v$ are fed into $f(v)$ to obtain $\hat{x}_{\theta}^+,\hat{x}_{\theta}^-, \hat{x}_{HR} $.
Then, by utilizing the $\mathcal{L}_{rec}$ from the SFT training phase, we derive new positive optimization objectives $\mathcal{L}_\theta^{'+}$ and negative optimization objectives $\mathcal{L}_\theta^{'-}$ in image and feature space:
\begin{equation}
    \mathcal{L}_{\theta}^{'+}=r\mathcal{L}_{Rec}(\hat{x}_{\theta}^+,\hat{x}_{HR}),
    \mathcal{L}_{\theta}^{'-}=(1-r)\mathcal{L}_{Rec}(\hat{x}_{\theta}^-,\hat{x}_{HR}).
\end{equation}
This results in the final training objective for DNF-SR as follows:
\begin{equation}
\begin{split}
    \mathcal{L}_{NF^2T}=\mathbb{E}_{\pi_{old}(\hat{z}_{HR}|z_{LR}),z_{LR}}  \Big[  r\mathcal{L}_{Rec}(\hat{x}_{\theta}^+,\hat{x}_{HR}) \\ + (1-r)\mathcal{L}_{Rec}(\hat{x}_{\theta}^-,\hat{x}_{HR}) \Big].
\end{split}
\label{eq:NF$^2$T_sr}
\end{equation}
Unlike reinforcement learning methods that require explicit probability modeling, NF$^2$T directly enhances model performance through implicit optimization directions. This characteristic makes it particularly well-suited for one-step SR tasks. Furthermore, in contrast to DiffusionDPO, NF$^2$T can leverage multiple sampled images to determine the optimization direction, whereas DiffusionDPO is limited to pairwise preference data, which restricts data utilization efficiency during training.

\noindent \textbf{Reward Calculation.}
Regarding the calculation of rewards, for the $K$ samples generated during each optimization step, we employ multiple evaluation metrics as rewards to ensure the accuracy of the optimization direction.
The evaluation metrics used are broadly categorized into full-reference and no-reference metrics.
Full-reference (FR) metrics include LPIPS \cite{blau2018perception} and DISTS \cite{ding2020image}, while no-reference (NR) metrics include CLIPIQA \cite{wang2023clipiqa}, MUSIQ \cite{ke2021musiq} and MANIQA \cite{yang2022maniqa}.
After computing $r_i^{raw}$ for the $i$-th metric, we standardize the rewards within the batch to obtain $r_i^{std}$.
Assuming $r_i^{std}$ follows a standard Gaussian distribution, we obtain the normalized reward $r_{i}$ by calculating $r_{i} = \Phi(X < r_i^{\text{std}})$,
where $\Phi(\cdot)$ is the standard Gaussian cumulative distribution function.
Finally, the $r_i$ values from different metrics are averaged to yield the ultimate reward.

\section{Experiments}

\subsection{Experimental Settings.}

\noindent \textbf{Training Datasets.}
Following DiT4SR \cite{duan2025dit4sr}, we adopt DIV2K ~\cite{agustsson2017ntire}, DIV8K \cite{gu2019div8k}, Flickr2K \cite{timofte2017ntire}, NKUSR8K \cite{duan2025dit4sr}, and the first 10K face images from FFHQ \cite{karras2019style} as our training dataset. Additionally, we utilize Real-ESRGAN's \cite{wang2021real} degradation pipeline to synthesize LR-HR paired data.

\noindent \textbf{Test Datasets.}
We evaluate our model using four datasets: RealSR \cite{cai2019toward}, DrealSR \cite{wei2020component}, DIV2K-Val \cite{agustsson2017ntire}, and RealLQ250 \cite{ai2025dreamclear}. Among these, DIV2K-Val is a synthetic dataset. RealSR and DrealSR are real-world datasets that include reference HR images, while RealLQ250 is a real-world dataset without reference HR images.

\noindent \textbf{Evaluation Metrics.}
We employ two reference-based evaluation metrics, PSNR and LPIPS~\cite{blau2018perception}, to assess image fidelity and perceptual quality, respectively.
However, as claimed in previous studies~\cite{jinjin2020pipal,yu2024scaling,duan2025dit4sr}, full-reference metrics, such as PSNR and SSIM~\cite{wang2004ssim}, often struggle to accurately reflect the visual effects of restored results.
Consequently, we employ several no-reference metrics, including MUSIQ\cite{ke2021musiq}, MANIQA~\cite{yang2022maniqa}, CLIPIQA~\cite{wang2023clipiqa}, QALIGN~\cite{wu2024qalign} and VQ-R1~\cite{wu2025vqr1}, to evaluate image quality.

\noindent \textbf{Implementation Details.}
We utilize FLUX.1-Kontext-dev~\cite{batifol2025flux} as our pre-trained model.
During the supervised fine-tuning stage, we fine-tune the VAE Encoder and DiT Block using LoRA~\cite{hu2022lora}, while keeping the VAE Decoder fixed. In Eq. \eqref{sft_loss}, we adopt the same settings in OMGSR \cite{wu2025omgsr}, where $\lambda_1=5$, $\lambda_2=2$, $\lambda_3=5$, $\lambda_4=0.5$.
We employ the AdamW~\cite{loshchilov2017decoupled} optimizer with a learning rate of 2e-5 and a batch size of 1, training for 5000 steps across 8 H20 GPUs.
In the post-training phase, we exclusively fine-tune the DiT Block with LoRA, setting the number of samples per optimization step to 8.
For both of these stages, we set the LoRA rank to 64.
We employ Qwen3-VL(8B) as an image caption generator to produce captions for both the training-stage high-resolution (HR) images and the inference-stage low-resolution (LR) images.
The same caption is also employed during inference for other models.

\subsection{Comparison with Existing Methods}

\noindent \textbf{Compared Methods.}
We compare our method with SOTA Diffusion-based Real-ISR methods, which include both single-step and multi-step super-resolution approaches.
The multi-step methods comprise DiffBIR~\cite{lin2024diffbir}, SeeSR~\cite{wu2024seesr}, and DiT4SR~\cite{duan2025dit4sr}, while the single-step methods include S3Diff~\cite{zhang2024degradation}, PisaSR~\cite{sun2024pisasr},  SinSR-1s~\cite{wang2024sinsr}, OSEDiff~\cite{wu2024osediff}, TSDSR~\cite{dong2025tsd}, HYPIR~\cite{lin2025hypir}, and OMGSR~\cite{wu2025omgsr}.

\begin{table}[htbp]
  \centering
  \caption{Ablation of the DNF-SR model design. $^*$ denotes the use of the text-to-image model Flux as the pre-trained model; all others use Flux-Kontext.}
  \resizebox{\linewidth}{!}{
    \begin{tabular}{lc|cccc}
        \toprule
        \textbf{Input} & \textbf{Timestep} & \textbf{LPIPS↓} & \textbf{MUSIQ↑} & \textbf{MANIQA↑} & \textbf{QALIGN↑} \\
        \midrule
        $\displaystyle z_{\text{mix}}^*$ & t=0.5 & 0.3349& 69.838 & 0.6729& \textcolor{red}{\textbf{3.9363}} \\
        $\displaystyle z_{\text{LR}}^*$ & t=0.5 & 0.3040& 70.441 & \textcolor{blue}{\underline{0.6808}}& 3.8771 \\
        $\displaystyle (z_{\text{mix}},z_{\text{LR}})^*$ & t=0.5 & \textcolor{blue}{\underline{0.2995}}& \textcolor{red}{\textbf{71.116}} & 0.6718& 3.9057 \\
       \rowcolor{gray!15} $\displaystyle (z_{\text{mix}},z_{\text{LR}})$ & t=0.5 & \textcolor{red}{\textbf{0.2925}}& \textcolor{blue}{\underline{70.672}} & \textcolor{red}{\textbf{0.6903}}& \textcolor{blue}{\underline{3.9162}} \\
        \midrule
        $\displaystyle (\epsilon,z_{\text{LR}})$ & t=1   & 0.30250 & 70.333 & 0.6853& 3.8574 \\
        $\displaystyle (z_{\text{mix}},z_{\text{LR}})$ & t=0.25 & \textcolor{blue}{\underline{0.2945}}& 70.493 & \textcolor{blue}{\underline{0.6870}}& 3.8688 \\
        $\displaystyle (z_{\text{mix}},z_{\text{LR}})$ & t=0.75 & 0.2960& \textcolor{blue}{\underline{70.526}} & 0.6790& \textcolor{red}{\textbf{3.9304}} \\
      \rowcolor{gray!15}  $\displaystyle (z_{\text{mix}},z_{\text{LR}})$ & t=0.5 & \textcolor{red}{\textbf{0.2925}}& \textcolor{red}{\textbf{70.672}} & \textcolor{red}{\textbf{0.6903}}& \textcolor{blue}{\underline{3.9162}} \\
        \bottomrule
    \end{tabular}
    }
  \label{tab:ab_sft}
\end{table}

\begin{figure}
    \centering
    \includegraphics[width=\linewidth]{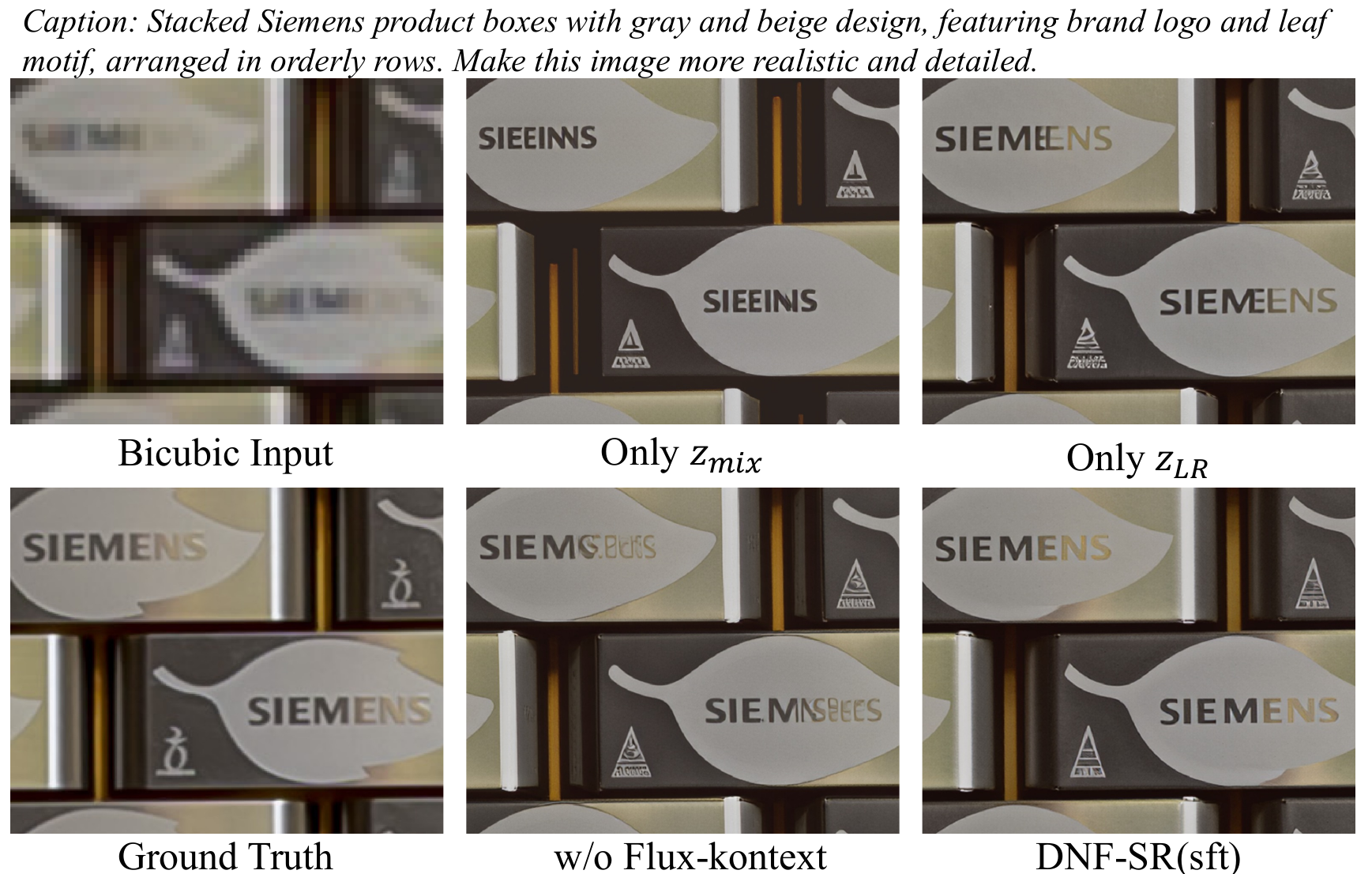}
    \caption{Visualizing the ablation study of the model design.}
    \label{fig:ab_sft}
\end{figure}

\noindent \textbf{Quantitative Comparisons.}
Quantitative comparisons with state-of-the-art Real-ISR methods on four benchmarks are presented in Tab.~\ref{tab:main_tab}.
It shows that our DNF-SR achieves overwhelming performance across all no-reference metrics on all benchmarks.
Even DNF-SR(sft), which only undergoes supervised fine-tuning without negative-aware feature fine-tuning, still obtains highly competitive performance.
Notably, we only use no-reference metrics CLIPIQA, MANIQA, and MUSIQ to calculate the reward, and do not employ the multimodal large language model-based evaluation methods QALIGN and VQ-R1.
The state-of-the-art evaluation scores of our method on QALIGN and VQ-R1 fully demonstrate its performance.

\noindent \textbf{Qualitative Comparisons.}
During inference, we feed the LR image into Qwen3-VL to generate captions, which are then applied to various super-resolution algorithms.
The qualitative comparison with other methods is presented in Fig.~\ref{fig:visual_all}.
From the results in the first row, it can be observed that DNF-SR can accurately generate ``NOKIA" even when the LR image content is severely corrupted. This is attributed to our improved utilization of the diffusion model's prior, allowing it to retain strong generative capabilities when applied to super-resolution tasks. It can be seen that DiT4SR also generates a relatively good logo, but due to its multi-step super-resolution approach, DNF-SR has an advantage in performance. The results from the second row indicate that our method can produce more realistic outcomes (e.g., the dog's toes and fur). The third row demonstrates that our method can generate clearer results.

\subsection{Ablation Study}
To further demonstrate the effectiveness of DNF-SR, we conduct an ablation study on RealSR. Specifically, we perform ablation experiments on the \textbf{model design} and post-training algorithm NF$^2$T used in DNF-SR, respectively.

\begin{figure}
    \centering
    \includegraphics[width=\linewidth]{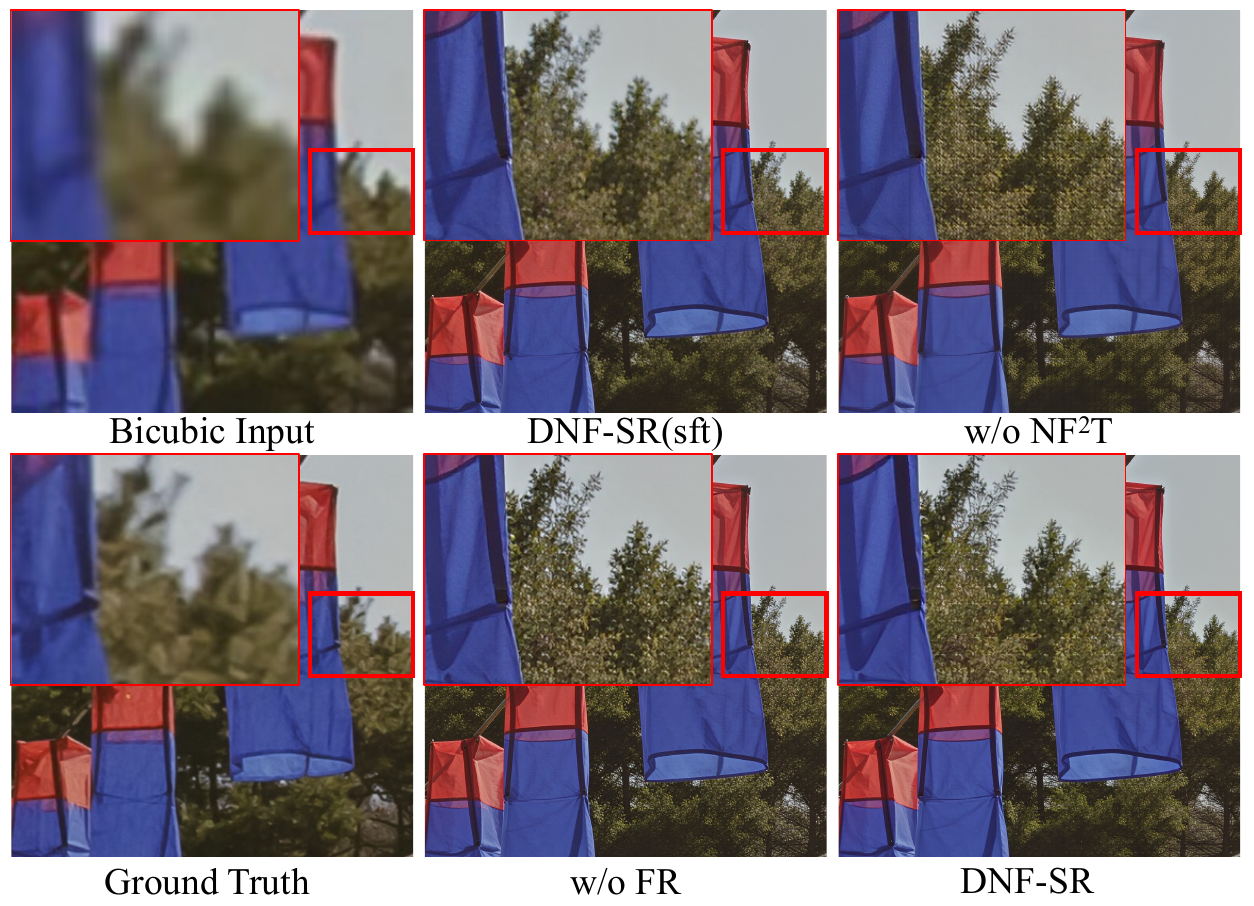}
    \caption{Visualizing the ablation study of the $NF^2T$ method.}
    \label{fig:ab_nf2t}
\end{figure}

\noindent \textbf{Model Design.}
In Tab.~\ref{tab:ab_sft}, we conduct ablation experiments on three aspects of the DNF-SR model structure: dual-path input, using the editing model, and denoising at intermediate timesteps.
1)\textbf{ dual-path input and using editing model.} Tab.~\ref{tab:ab_sft} shows that the dual-input in DNF-SR achieves better performance compared to using only LR latent $z_{LR}$or only noisy LR latent $z_{mix}$. Additionally, using the image editing model FLUX-Kontext further enhances performance. As depicted in Fig.~\ref{fig:ab_nf2t}, DNF-SR effectively leverages the prior knowledge of the generative model under a given caption, producing more accurate content while maintaining fidelity.
2) \textbf{denoising at intermediate timesteps.} We also report the metrics for single-step denoising of DNF-SR at different timesteps $t$.
Compared to denoising from pure noise in a single step, denoising at any intermediate timestep yields better performance. To balance realism and fidelity, we chose $t=0.5$.

\noindent \textbf{Negative-aware Feature Fine-Turing.}
In Tab. \ref{tab:ab_NF2T}, we conduct ablation experiments on the post-training algorithm of our proposed NF$^2$T.
It can be observed that DNF-SR achieves significant improvements in all no-reference metrics, especially QALIGN (not involved in reward calculation).
As shown in Fig. \ref{fig:ab_nf2t}, DNF-SR with NF$^2$T yields notably clearer results. In contrast, the HR ground truth is blurrier, which leads to a decrease in the reference-based LPIPS metric for DNF-SR.
Meanwhile, as shown in Fig. \ref{fig:ab_nf2t}, if NFT is used for optimization in the latent space, it exhibits obvious grid artifacts. Regarding the use of the reward model, we find that incorporating full-reference (FR) metrics when using NF2T can further improve no-reference (NR) metrics while preventing degradation of FR metrics. Additionally, it can generate more natural results.

\begin{table}[htbp]
  \centering
  \caption{Ablation study of the NF$^2$T method.}
  \resizebox{\linewidth}{!}{
    \begin{tabular}{lc|cccc}
    \toprule
    \textbf{Setting} & \textbf{Reward Model} & \textbf{LPIPS↓} & \textbf{MUSIQ↑} & \textbf{MANIQA↑} & \textbf{QALIGN↑} \\
    \midrule
    -      &   -    & \textcolor[rgb]{ 1,  0,  0}{\textbf{0.2925}} & 70.6717 & 0.6903 & 3.9162 \\
    NFT   & NR+FR & 0.3250 & \underline{\textcolor{blue}{71.1682}} & 0.6413 & 3.943 \\
    NF$^2$T  & NR    & 0.3255 & 71.0911 & \underline{\textcolor{blue}{0.6925}} & \underline{\textcolor{blue}{3.9979}} \\
   \rowcolor{gray!15}  NF$^2$T  & NR+FR & \underline{\textcolor{blue}{0.3239}} & \textcolor[rgb]{ 1,  0,  0}{\textbf{72.0396}} & \textcolor[rgb]{ 1,  0,  0}{\textbf{0.693}} & \textcolor[rgb]{ 1,  0,  0}{\textbf{4.0718}} \\
    \bottomrule
    \end{tabular}
    }
  \label{tab:ab_NF2T}
\end{table}

\section{Conclusion}
This paper presents DNF-SR, a novel framework for real-world image super-resolution (Real-ISR). DNF employs a dual-input architecture that concatenates noisy LR latents with original LR latents to narrow the distribution gap between LR inputs and diffusion models' native inputs while preserving content fidelity. Additionally, the Flux-Kontext image-editing pretrained model is utilized to better leverage generative model priors. Furthermore, we propose Negative-aware Feature Fine-Tuning (NF²T), which shifts optimization from the latent space to image and feature spaces, using aggregated rewards to enhance realism and suppress artifacts. Experiments demonstrate that DNF-SR achieves state-of-the-art performance across multiple benchmarks and can generate superior results.

\section{Acknowledgments}
This work was supported in part by the National Natural Science Foundation of China (62306153, 62225604), Tianjin Natural Science Foundation Project (25ZXRGGX00290, 24JCJQJC00020, 25JCQNJC01390), the Young Elite Scientists Sponsorship Program by CAST (YESS20240686), the Fundamental Research Funds for the Central Universities (Nankai University, 63253223, 63253219), ``Science and Technology Yongjiang 203" key technology breakthrough plan project (2024Z120), Chinese government-guided local science and technology development fund projects (scientific and technological achievement transfer and transformation projects)(254Z0102G) and Shenzhen Science and Technology Program (JCYJ20240813114237048).
The computational devices is supported by the Supercomputing Center of Nankai University (NKSC).

\clearpage

\appendix

\section*{Supplementary Material}

\begin{table*}[!t]
    \centering
    \caption{A comprehensive evaluation against state-of-the-art GAN-based methods across synthetic and real-world datasets. The top-performing results under each metric are marked in \textcolor[rgb]{1,0,0}{\textbf{red}}.}
    \label{tab:gan_comparison}
    \resizebox{\textwidth}{!}{
    \small
    \begin{tabular}{c|c|ccccccc}
        \toprule
        Datasets & Methods & PSNR $\uparrow$ & LPIPS~\cite{blau2018perception} $\downarrow$ & CLIPIQA~\cite{wang2023clipiqa} $\uparrow$ & MUSIQ~\cite{ke2021musiq} $\uparrow$ & MAINIQA~\cite{yang2022maniqa} $\uparrow$ & QALIGN~\cite{wu2024qalign} $\uparrow$ & VQ-R1~\cite{wu2025vqr1} $\uparrow$ \\
        \midrule
        \multirow{4}{*}{\textit{DIV2k}}   & BSRGAN    & \textcolor[rgb]{1,0,0}{\textbf{24.583}} & 0.3351 & 0.5246 & 61.193 & 0.5040 & 3.1703 & 3.3063 \\
                                          & RealESRGAN& 24.293 & \textcolor[rgb]{1,0,0}{\textbf{0.3112}} & 0.5276 & 61.049 & 0.5484 & 3.2764 & 3.2623 \\
                                          & LDL       & 23.828 & 0.3256 & 0.5179 & 60.040 & 0.5328 & 3.1798 & 3.1018 \\
                                          & DNF-SR    & 23.631 & 0.3234 & \textcolor[rgb]{1,0,0}{\textbf{0.7723}} & \textcolor[rgb]{1,0,0}{\textbf{71.546}} & \textcolor[rgb]{1,0,0}{\textbf{0.6703}} & \textcolor[rgb]{1,0,0}{\textbf{4.1563}} & \textcolor[rgb]{1,0,0}{\textbf{4.3630}} \\
        \midrule
        \multirow{4}{*}{\textit{DrealSR}} & BSRGAN    & \textcolor[rgb]{1,0,0}{\textbf{28.702}} & 0.2858 & 0.5092 & 57.159 & 0.4844 & 2.9572 & 3.0559 \\
                                          & RealESRGAN& 28.618 & 0.2818 & 0.4517 & 54.275 & 0.4902 & 2.8638 & 2.7683 \\
                                          & LDL       & 28.196 & \textcolor[rgb]{1,0,0}{\textbf{0.2790}} & 0.4473 & 53.948 & 0.4894 & 2.8576 & 2.6129 \\
                                          & DNF-SR    & 28.141 & 0.3531 & \textcolor[rgb]{1,0,0}{\textbf{0.7559}} & \textcolor[rgb]{1,0,0}{\textbf{68.732}} & \textcolor[rgb]{1,0,0}{\textbf{0.6515}} & \textcolor[rgb]{1,0,0}{\textbf{3.7997}} & \textcolor[rgb]{1,0,0}{\textbf{3.9152}} \\
        \midrule
        \multirow{4}{*}{\textit{RealSR}}  & BSRGAN    & \textcolor[rgb]{1,0,0}{\textbf{26.379}} & \textcolor[rgb]{1,0,0}{\textbf{0.2656}} & 0.5114 & 63.283 & 0.5419 & 3.1829 & 3.4907 \\
                                          & RealESRGAN& 25.687 & 0.2710 & 0.4489 & 60.364 & 0.5504 & 3.1081 & 3.1342 \\
                                          & LDL       & 25.280 & 0.2750 & 0.4556 & 60.930 & 0.5495 & 3.0898 & 2.9897 \\
                                          & DNF-SR    & 24.970 & 0.3239 & \textcolor[rgb]{1,0,0}{\textbf{0.7257}} & \textcolor[rgb]{1,0,0}{\textbf{70.040}} & \textcolor[rgb]{1,0,0}{\textbf{0.6930}} & \textcolor[rgb]{1,0,0}{\textbf{4.0718}} & \textcolor[rgb]{1,0,0}{\textbf{4.2646}} \\
        \midrule
        \multirow{4}{*}{\textit{RealLQ250}}& BSRGAN   & - & - & 0.5940 & 66.289 & 0.5963 & 3.4794 & 3.8124 \\
                                           & RealESRGAN& - & - & 0.6253 & 66.990 & 0.6148 & 3.6471 & 3.8088 \\
                                           & LDL      & - & - & 0.6183 & 67.027 & 0.6147 & 3.6357 & 3.6940 \\
                                           & DNF-SR   & - & - & \textcolor[rgb]{1,0,0}{\textbf{0.7997}} & \textcolor[rgb]{1,0,0}{\textbf{73.700}} & \textcolor[rgb]{1,0,0}{\textbf{0.7029}} & \textcolor[rgb]{1,0,0}{\textbf{4.4752}} & \textcolor[rgb]{1,0,0}{\textbf{4.6090}} \\
        \bottomrule
    \end{tabular}
    }
\end{table*}
In this Supplementary Material, we provide additional details, including the comparison with GAN-based methods in Section~\ref{app:gan}, discussion on perception and restoration in Section~\ref{app:discuss}, more visual comparisons in Section~\ref{app:visual_more}, and the algorithm in Section~\ref{app:algorithm}. We conduct these additional comparisons and analyses to validate the effectiveness of DNF-SR.

\section{Comparison with GAN-based Methods}
\label{app:gan}
We compare DNF-SR with three GAN-based Real-ISR methods: BSRGAN \cite{zhang2021designing}, RealESRGAN \cite{wang2021real}, and LDL \cite{liang2022details}.
Quantitative evaluations are conducted on the DIV2K \cite{agustsson2017ntire}, RealSR \cite{cai2019toward},  DrealSR \cite{wei2020component} and RealLQ \cite{ai2025dreamclear} datasets, with results summarized in Tab.~\ref{tab:gan_comparison}.
The experimental results demonstrate that DNF-SR, leveraging a dual-input strategy and a novel post-training optimization method NF²T, achieves significantly superior no-reference metrics compared to GAN-based methods.

\begin{figure*}
    \centering
    \includegraphics[height=\textheight - 22pt,width=\linewidth]{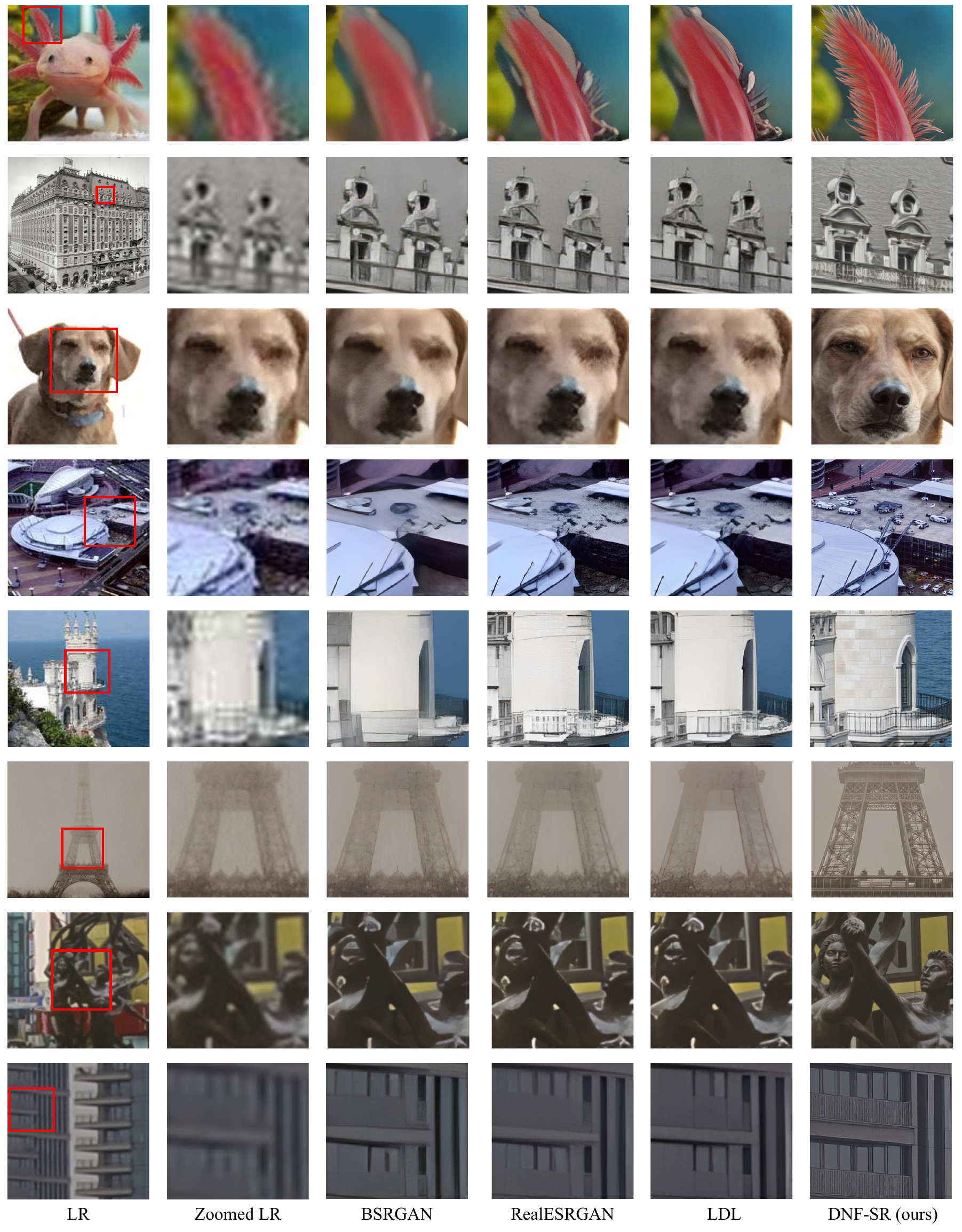}
    \caption{Vision comparisons between DNF-SR and GAN-based Real-ISR methods~\cite{wang2021real,zhang2021designing, liang2022details}. Zoom in for a better view.}
    \label{fig:gan_vis}
\end{figure*}

Additionally, Fig~\ref{fig:gan_vis} presents a visual comparison between DNF-SR and other GAN-based methods.
The results show that DNF-SR reconstructs more photorealistic and natural outcomes.
When compared to GAN-based methods, DNF-SR demonstrates distinct advantages in visual fidelity.
Specifically, it achieves higher precision in restoring structured elements (e.g., text and architectural details) while rendering complex materials such as fabrics and natural textures with enhanced realism.
This enables DNF-SR to more accurately reproduce the fine-grained detail hierarchy and authentic visual texture characteristic of high-resolution images, outperforming the GAN-based methods in both structural integrity and perceptual quality.

\begin{figure}
    \centering
    \includegraphics[width=\linewidth]{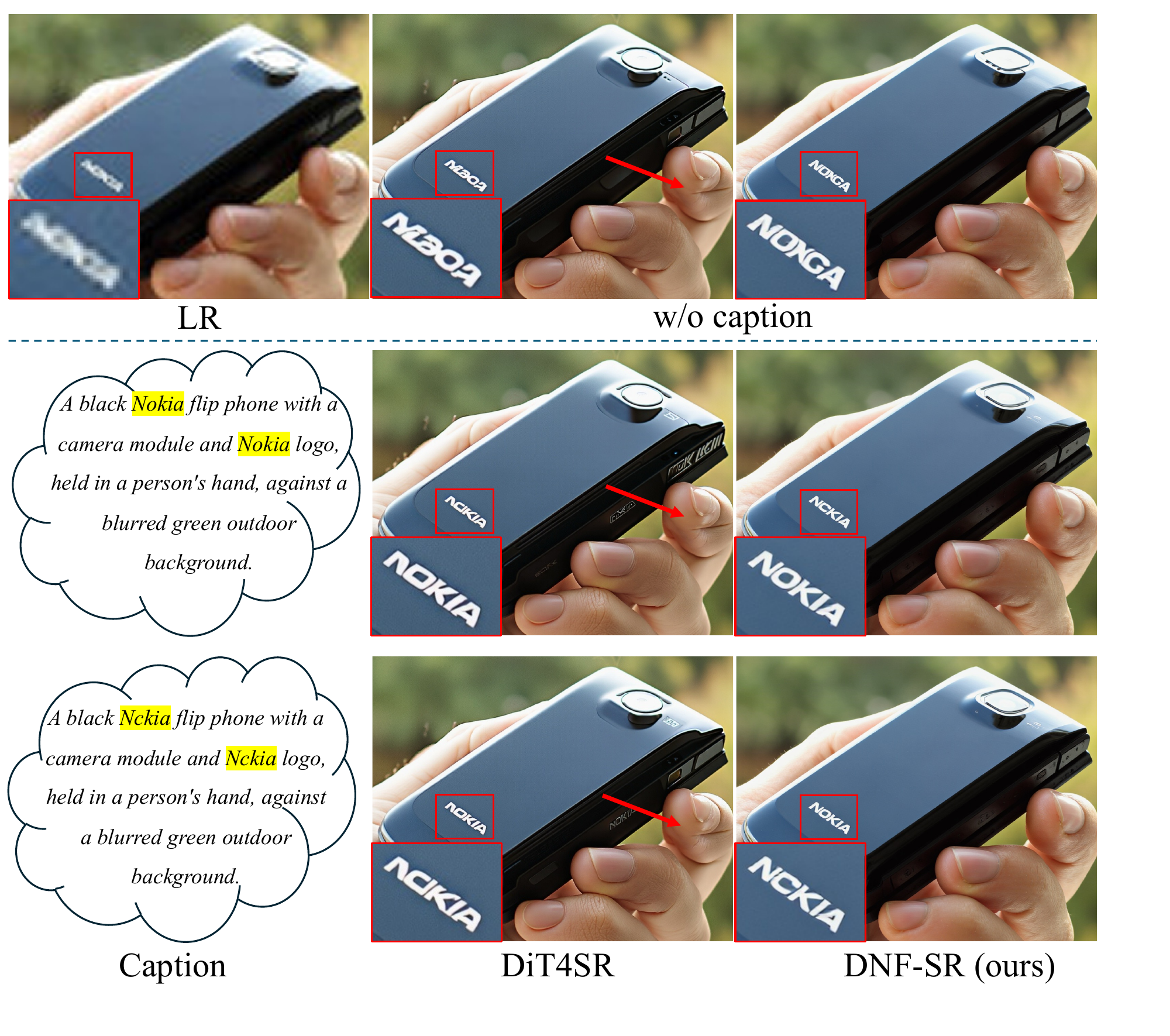}
    \caption{Visual comparisons between DNF-SR and DiT4SR when using different captions including no caption, a reasonable caption with ``Nokia", and an unreasonable caption where ``Nokia" is replaced with ``Nckia". DiT4SR exhibits structural issues at the position indicated by the red arrow.}
    \label{fig:caption}
\end{figure}

\section{Discussion on Perception and Restoration}
\label{app:discuss}
In experiments, we observe that existing multi-modal large language models (MLLMs) can effectively perceive the content in images.
Even for challenging low-resolution (LR) images, they can infer reasonable content for blurred regions based on the overall image information.
Meanwhile, current DiT-based generative models can adhere well to captions for image generation.
However, in SR tasks, restoring strongly semantic structures such as text and logos is extremely challenging.
As shown in the first row of Fig.~\ref{fig:caption}, when no additional caption is used for the SR model, it is difficult to restore text with normal semantics.
As shown in Fig~\ref{fig:caption}, when captions generated by MLLMs are used as conditions, DiT-based SR models can effectively alleviate this issue.
Nevertheless, when we manually replace ``Nokia'' with ``Nckia'', DiT4SR exhibits poor prompt-following performance.
This is because when applying text-to-image models to SR task, the model is caused to focus more on LR images, which impairs the inherent prompt-following capability of the original text-to-image model.
In contrast, our method DNF-SR narrows the gap between LR and the original input of generative models through a dual-path input design.
It also initializes with an image editing model to better perceive the information of LR used as conditions.
This enables DNF-SR to better preserve the inherent generation capability and prompt-following ability of the original generative model when applied to SR tasks, thereby allowing it to better adhere to captions and restore more realistic and reasonable images.

However, current SR methods still obtain captions by leveraging MLLMs to perceive image content before applying them to restoration tasks.
Only using text caption may sometimes fail to fully convey image information, and the separate execution of perception and restoration steps also introduces redundancy.
Thus, it is highly meaningful to develop an integrated perception-restoration model that can accurately perceive semantic information in LR images and perform restoration within a single framework.

\section{More Visual Comparisons}
\label{app:visual_more}

\begin{figure*}
    \centering
    \includegraphics[height=\textheight - 22pt,width=\linewidth]{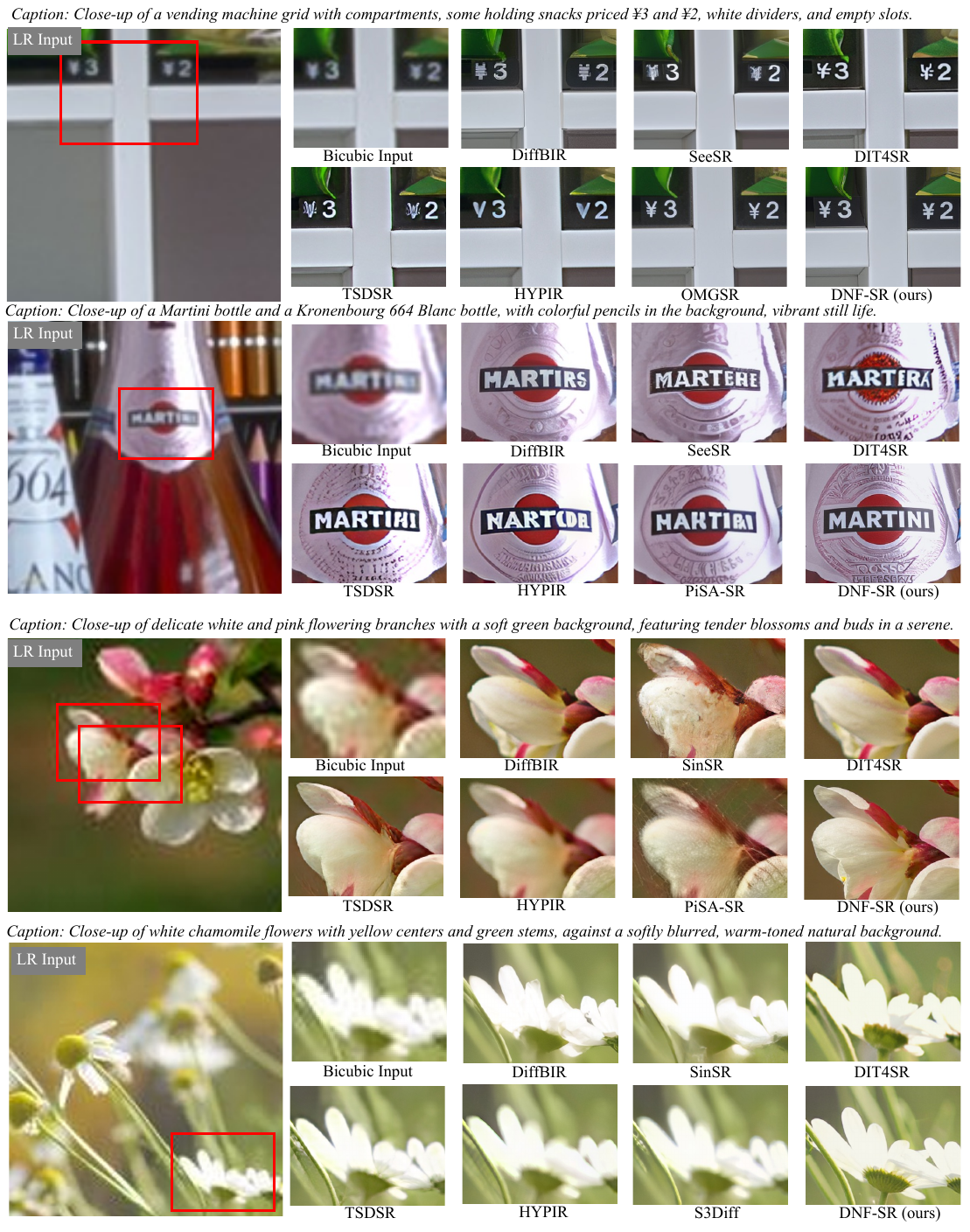}
    \caption{Vision comparisons between DNF-SR and different diffusion-based Real-ISR methods~\cite{lin2024diffbir, wu2024seesr, duan2025dit4sr, dong2025tsd, lin2025hypir, wu2025omgsr, zhang2024degradation, wang2024sinsr, sun2024pisasr}.}
    \label{fig:supp_visual1}
\end{figure*}

\begin{figure*}
    \centering
    \includegraphics[height=\textheight - 22pt,width=\linewidth]{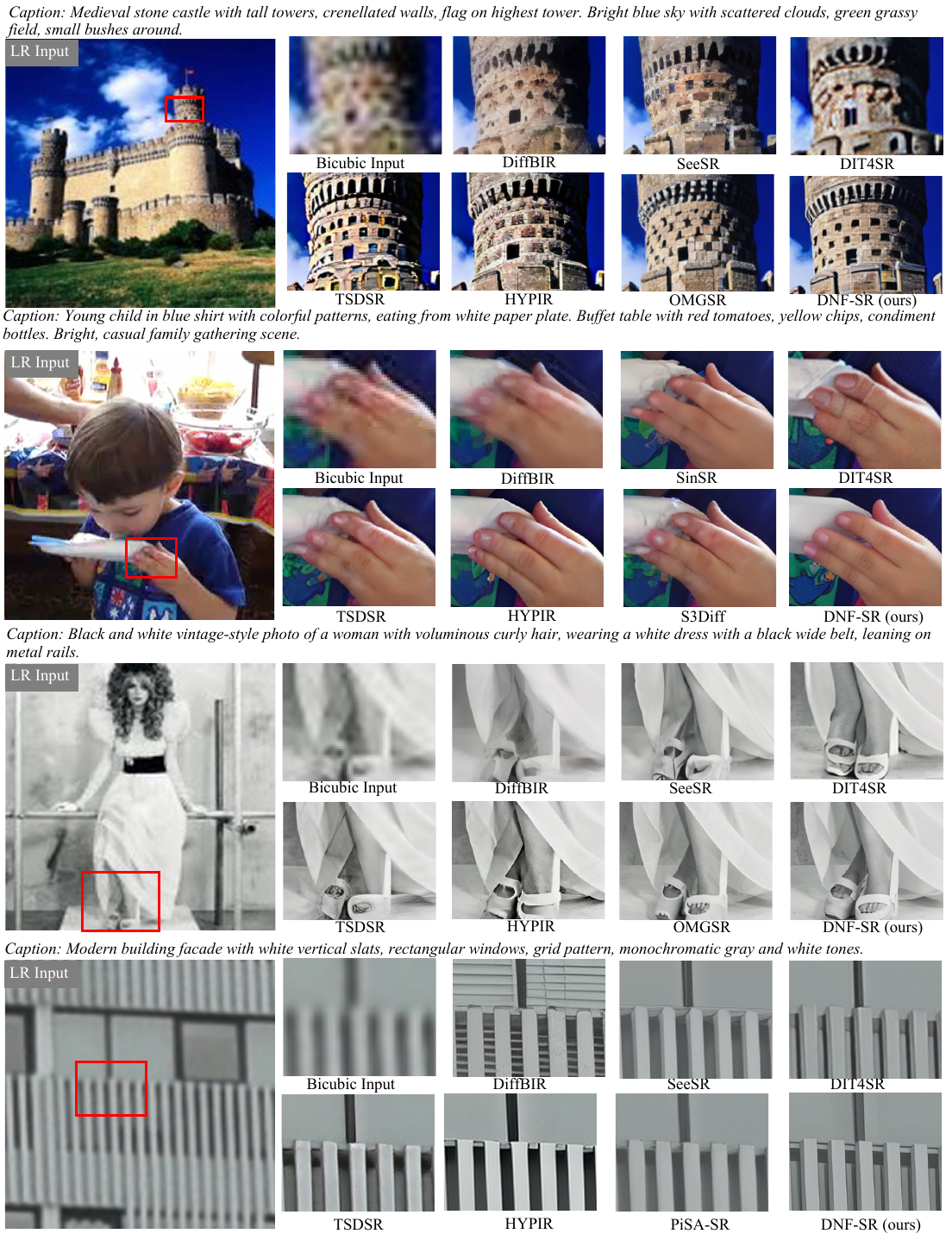}
    \caption{Vision comparisons between DNF-SR and different diffusion-based Real-ISR methods ~\cite{lin2024diffbir, wu2024seesr, duan2025dit4sr, dong2025tsd, lin2025hypir, wu2025omgsr, zhang2024degradation, wang2024sinsr, sun2024pisasr}.}
    \label{fig:supp_visual2}
\end{figure*}

In Fig.~\ref{fig:supp_visual1} and \ref{fig:supp_visual2}, we provide more visual comparisons with other diffusion-based Real-ISR methods.
As shown in Fig.~\ref{fig:supp_visual1}, DNF-SR can better adhere to the content of the caption and restore more accurate text information.
And in close-up scenarios, DNF-SR can better restore the texture and details of the image.
Meanwhile, as shown in Fig.~\ref{fig:supp_visual2}, DNF-SR can restore more realistic images under severe degradation.
These examples all demonstrate the performance and robustness of DNF-SR for Real-ISR.

\section{Algorithm Details}
\label{app:algorithm}

The training of DNF-SR consists of two stages: supervised fine-tuning (SFT) and post-training.
In the SFT stage, we use multiple losses to perform supervised fine-tuning on the paired $(x_L, x_H, c)$ dataset.
In the post-training stage, we adopt a Negative-aware Feature Fine-Tuning method for reinforcement learning.
Specifically, we sample $K$ noises to generate $K$ restored images, then compute rewards using multiple reward functions, which are normalized and aggregated into a single $r$. Subsequently, we define positive and negative optimization directions to improve model performance. Here, \(K = 8\). Details are in Algorithm ~\ref{ap:algo}.

\begin{algorithm*}[t]
\caption{Training Procedure of Negativate-aware Feature Fine-tuning in DNF-SR}
\SetKwInput{KwInput}{Input}
\SetKwInput{KwOutput}{Output}
\DontPrintSemicolon
\label{ap:algo}
  \KwInput{Training datasets $\{x_L,x_H,c\}$ , fine-tuning one-step Diffusion-based SR Model including VAE encoder $E_{ref}$ and velocity prediction network $v_{ref}$, pre-trained VAE decoder $D_\varphi$, number of samples $K$ per training step, $N$ raw reward functions $r^{raw}(\cdot) \in \mathbb{R}$, one fixed mid-timestep $t_{mid}$.}
  \KwOutput{Post-trained one-step velocity prediction network $v_{\theta}$ for SR.}
    {Initialize data collection policy for velocity prediction $v_{old} \leftarrow v_{ref} $.
     Initialize training policy for velocity prediction $v_{\theta} \leftarrow v_{ref} $.
     Initialize data buffer $\mathcal{D} \leftarrow \emptyset$}
\While{train}{
    \tcc{Rollout Step, Data Collection}
    \For{ $\text{each sampled data} (x_L,x_H,c) \sim \mathcal{D}$}{
        Sample $K$ standard Gaussian noises $\epsilon^{1:K}$ and collect $K$ restored images $\hat{x}_{H}^{1:K}$ using $v_{old}$.
        Compute rewards $\{r^{raw}_{1:N}\}^{1:K}$ using $N$ raw reward functions, respectively.
        Standardize raw rewards in group: $r_i^{std} := (r_i^{raw} - mean(\{r^{raw}_{i}\}^{1:K})) / std(\{r^{raw}_{i}\}^{1:K})$.
        Normalize rewards using the standard Gaussian cumulative distribution function: $r_{i}=\Phi(X <r^{std})$.
        Average the $N$ rewards: $r = avg(r_{1:N})$
        $D \leftarrow \{c, x_{L}, \hat{x}_{H}^{1:K}, r^{1:K}\}$
    }
    \tcc{Gradient Step, Policy Optimization}
    \For{ each mini batch $\{ {c,x_{L}, \hat{x}_{H}, r}\}$}{
        Encode the LR image: $z_L = E_{ref}(x_{L})$.
        Forward diffusion process: $z_t = t_{mid}\epsilon+(1-t_{mid})z_{L}$.
        \tcc{Calculate Positive Optimization Direction}
        Implicit positive velocity: $v_\theta^+:= (1-\beta)v^{old}(z_{t},c,t_{mid}) + \beta v_\theta(z_{t},c, t_{mid})$.
        Implicit positive image: $\hat{x}_\theta^+:= D_\varphi(z_t-t_{mid}v_\theta^+)$.
        Positive optimization direction: $\mathcal{L}_\theta^+=r\mathcal{L}_{rec}(\hat{x}_\theta^+,\hat{x}_H)$.
        \tcc{Calculate Positive Optimization Direction}
        Implicit negative velocity: $v_\theta^-:= (1+\beta)v^{old}(z_{t},c,t_{mid}) - \beta v_\theta(z_{t},c, t_{mid})$.
        Implicit negative image: $\hat{x}_\theta^-:= D_\varphi(z_t-t_{mid}v_\theta^-)$.
        Negative optimization direction: $\mathcal{L}_\theta^-=(1-r)\mathcal{L}_{rec}(\hat{x}_\theta^-,\hat{x}_H)$.
        \tcc{Update Model Parameters}
        $\theta \leftarrow \theta - \lambda \nabla_\theta \left[ \mathcal{L}_\theta^+ + \mathcal{L}_\theta^- \right]$
    }
    \tcc{Online Update}
    Update data collection policy $v_{old} \leftarrow v_\theta$, and clear buffer $\mathcal{D} \leftarrow \emptyset.$
  }
\end{algorithm*}
{
    \small
    \bibliographystyle{ieeenat_fullname}
    \bibliography{main}

@String(CVPR= {IEEE Conf. Comput. Vis. Pattern Recog.})

@String(ICCV= {Int. Conf. Comput. Vis.})

@String(ECCV= {Eur. Conf. Comput. Vis.})

@String(TIP  = {IEEE Trans. Image Process.})

@String(ICLR = {Int. Conf. Learn. Represent.})

@String(AAAI = {AAAI})

@String(CVPRW= {IEEE Conf. Comput. Vis. Pattern Recog. Worksh.})

@String(CVPR  = {CVPR})

@String(ICCV  = {ICCV})

@String(ICML  = {ICML})

@String(ECCV  = {ECCV})

@String(TIP   = {IEEE TIP})

@String(ICLR  = {ICLR})

@String(CVPRW= {CVPRW})

@String(ICCVW= {ICCVW})

@inproceedings{goodfellow2020generative,
  title={Generative adversarial nets},
  author={Goodfellow, Ian J and Pouget-Abadie, Jean and Mirza, Mehdi and Xu, Bing and Warde-Farley, David and Ozair, Sherjil and Courville, Aaron and Bengio, Yoshua},
  booktitle={NeurIPS},
  year={2014}
}

@inproceedings{wang2021real,
  title={Real-esrgan: Training real-world blind super-resolution with pure synthetic data},
  author={Wang, Xintao and Xie, Liangbin and Dong, Chao and Shan, Ying},
  booktitle=ICCV,
  year={2021}
}

@inproceedings{liang2022details,
  title={Details or artifacts: A locally discriminative learning approach to realistic image super-resolution},
  author={Liang, Jie and Zeng, Hui and Zhang, Lei},
  booktitle=CVPR,
  year={2022}
}

@article{ding2020image,
  title={Image quality assessment: Unifying structure and texture similarity},
  author={Ding, Keyan and Ma, Kede and Wang, Shiqi and Simoncelli, Eero P},
  journal={TPAMI},
  year={2020}
}

@inproceedings{liuflow,
  title={Flow Straight and Fast: Learning to Generate and Transfer Data with Rectified Flow},
  author={Liu, Xingchao and Gong, Chengyue and others},
  booktitle={ICLR},
  year={2023}
}

@inproceedings{lipman2023flow,
  title={Flow Matching for Generative Modeling},
  author={Lipman, Yaron and Chen, Ricky TQ and Ben-Hamu, Heli and Nickel, Maximilian and Le, Matt},
  booktitle={ICLR},
  year={2023}
}

@article{li2025fluxsr,
  title={One Diffusion Step to Real-World Super-Resolution via Flow Trajectory Distillation},
  author={Li, Jianze and Cao, Jiezhang and Guo, Yong and Li, Wenbo and Zhang, Yulun},
  journal={ArXiv preprint},
  year={2025}
}

@inproceedings{dong2025tsd,
  title={Tsd-sr: One-step diffusion with target score distillation for real-world image super-resolution},
  author={Dong, Linwei and Fan, Qingnan and Guo, Yihong and Wang, Zhonghao and Zhang, Qi and Chen, Jinwei and Luo, Yawei and Zou, Changqing},
  booktitle=CVPR,
  year={2025}
}

@inproceedings{lin2024diffbir,
  title={Diffbir: Toward blind image restoration with generative diffusion prior},
  author={Lin, Xinqi and He, Jingwen and Chen, Ziyan and Lyu, Zhaoyang and Dai, Bo and Yu, Fanghua and Qiao, Yu and Ouyang, Wanli and Dong, Chao},
  booktitle=ECCV,
  year={2024},
}

@article{zhang2024degradation,
  title={Degradation-guided one-step image super-resolution with diffusion priors},
  author={Zhang, Aiping and Yue, Zongsheng and Pei, Renjing and Ren, Wenqi and Cao, Xiaochun},
  journal={ArXiv preprint},
  year={2024}
}

@inproceedings{sun2025pixel,
  title={Pixel-level and semantic-level adjustable super-resolution: A dual-lora approach},
  author={Sun, Lingchen and Wu, Rongyuan and Ma, Zhiyuan and Liu, Shuaizheng and Yi, Qiaosi and Zhang, Lei},
  booktitle={CVPR},
  year={2025}
}

@inproceedings{yu2024scaling,
  title={Scaling up to excellence: Practicing model scaling for photo-realistic image restoration in the wild},
  author={Yu, Fanghua and Gu, Jinjin and Li, Zheyuan and Hu, Jinfan and Kong, Xiangtao and Wang, Xintao and He, Jingwen and Qiao, Yu and Dong, Chao},
  booktitle=CVPR,
  year={2024}
}

@inproceedings{wang2024sinsr,
  title={SinSR: diffusion-based image super-resolution in a single step},
  author={Wang, Yufei and Yang, Wenhan and Chen, Xinyuan and Wang, Yaohui and Guo, Lanqing and Chau, Lap-Pui and Liu, Ziwei and Qiao, Yu and Kot, Alex C and Wen, Bihan},
  booktitle=CVPR,
  year={2024}
}

@inproceedings{yue2024resshift,
  title={Resshift: Efficient diffusion model for image super-resolution by residual shifting},
  author={Yue, Zongsheng and Wang, Jianyi and Loy, Chen Change},
  booktitle={NeurIPS},
  year={2024}
}

@article{wu2024osediff,
  title={One-Step Effective Diffusion Network for Real-World Image Super-Resolution},
  author={Wu, Rongyuan and Sun, Lingchen and Ma, Zhiyuan and Zhang, Lei},
  journal={ArXiv preprint},
  year={2024}
}

@inproceedings{ai2025dreamclear,
  title={DreamClear: High-Capacity Real-World Image Restoration with Privacy-Safe Dataset Curation},
  author={Ai, Yuang and Zhou, Xiaoqiang and Huang, Huaibo and Han, Xiaotian and Chen, Zhengyu and You, Quanzeng and Yang, Hongxia},
  booktitle={NeurIPS},
  year={2025}
}

@article{xie2024addsr,
  title={Addsr: Accelerating diffusion-based blind super-resolution with adversarial diffusion distillation},
  author={Xie, Rui and Zhao, Chen and Zhang, Kai and Zhang, Zhenyu and Zhou, Jun and Yang, Jian and Tai, Ying},
  journal={ArXiv preprint},
  year={2024}
}

@inproceedings{wu2024seesr,
  title={Seesr: Towards semantics-aware real-world image super-resolution},
  author={Wu, Rongyuan and Yang, Tao and Sun, Lingchen and Zhang, Zhengqiang and Li, Shuai and Zhang, Lei},
  booktitle=CVPR,
  year={2024}
}

@article{podell2023sdxl,
  title={Sdxl: Improving latent diffusion models for high-resolution image synthesis},
  author={Podell, Dustin and English, Zion and Lacey, Kyle and Blattmann, Andreas and Dockhorn, Tim and M{\"u}ller, Jonas and Penna, Joe and Rombach, Robin},
  journal={ArXiv preprint},
  year={2023}
}

@article{chen2023pixart,
  title={Pixart-$alpha$: Fast training of diffusion transformer for photorealistic text-to-image synthesis},
  author={Chen, Junsong and Yu, Jincheng and Ge, Chongjian and Yao, Lewei and Xie, Enze and Wu, Yue and Wang, Zhongdao and Kwok, James and Luo, Ping and Lu, Huchuan and others},
  journal={ArXiv preprint},
  year={2023}
}

@inproceedings{rombach2022high,
  title={High-resolution image synthesis with latent diffusion models},
  author={Rombach, Robin and Blattmann, Andreas and Lorenz, Dominik and Esser, Patrick and Ommer, Bj{\"o}rn},
  booktitle=CVPR,
  year={2022}
}

@inproceedings{peebles2023scalable,
  title={Scalable diffusion models with transformers},
  author={Peebles, William and Xie, Saining},
  booktitle={ICCV},
  year={2023}
}

@misc{blackforestlabs2024,
  author={blackforestlabs.ai},
  title={Flux, offering state-of-the-art performance image generation},
  year={2024},
  url={https://blackforestlabs.ai/},
}

@inproceedings{zhang2023adding,
  title={Adding conditional control to text-to-image diffusion models},
  author={Zhang, Lvmin and Rao, Anyi and Agrawala, Maneesh},
  booktitle=CVPR,
  year={2023}
}

@inproceedings{saharia2022photorealistic,
  title={Photorealistic text-to-image diffusion models with deep language understanding},
  author={Saharia, Chitwan and Chan, William and Saxena, Saurabh and Li, Lala and Whang, Jay and Denton, Emily L and Ghasemipour, Kamyar and Gontijo Lopes, Raphael and Karagol Ayan, Burcu and Salimans, Tim and others},
  booktitle={NeurIPS},
  year={2022}
}

@article{liu2025flow,
  title={Flow-grpo: Training flow matching models via online rl},
  author={Liu, Jie and Liu, Gongye and Liang, Jiajun and Li, Yangguang and Liu, Jiaheng and Wang, Xintao and Wan, Pengfei and Zhang, Di and Ouyang, Wanli},
  journal={ArXiv preprint},
  year={2025}
}

@inproceedings{fan2023dpok,
  title={Dpok: Reinforcement learning for fine-tuning text-to-image diffusion models},
  author={Fan, Ying and Watkins, Olivia and Du, Yuqing and Liu, Hao and Ryu, Moonkyung and Boutilier, Craig and Abbeel, Pieter and Ghavamzadeh, Mohammad and Lee, Kangwook and Lee, Kimin},
  booktitle={NeurIPS},
  year={2023}
}

@article{black2023training,
  title={Training diffusion models with reinforcement learning},
  author={Black, Kevin and Janner, Michael and Du, Yilun and Kostrikov, Ilya and Levine, Sergey},
  journal={ArXiv preprint},
  year={2023}
}

@article{guo2025deepseek,
  title={Deepseek-r1: Incentivizing reasoning capability in llms via reinforcement learning},
  author={Guo, Daya and Yang, Dejian and Zhang, Haowei and Song, Junxiao and Zhang, Ruoyu and Xu, Runxin and Zhu, Qihao and Ma, Shirong and Wang, Peiyi and Bi, Xiao and others},
  journal={ArXiv preprint},
  year={2025}
}

@article{bai2022training,
  title={Training a helpful and harmless assistant with reinforcement learning from human feedback},
  author={Bai, Yuntao and Jones, Andy and Ndousse, Kamal and Askell, Amanda and Chen, Anna and DasSarma, Nova and Drain, Dawn and Fort, Stanislav and Ganguli, Deep and Henighan, Tom and others},
  journal={ArXiv preprint},
  year={2022}
}

@inproceedings{ouyang2022training,
  title={Training language models to follow instructions with human feedback},
  author={Ouyang, Long and Wu, Jeffrey and Jiang, Xu and Almeida, Diogo and Wainwright, Carroll and Mishkin, Pamela and Zhang, Chong and Agarwal, Sandhini and Slama, Katarina and Ray, Alex and others},
  booktitle={NeurIPS},
  year={2022}
}

@inproceedings{agustsson2017ntire,
  title={Ntire 2017 challenge on single image super-resolution: Dataset and study},
  author={Agustsson, Eirikur and Timofte, Radu},
  booktitle=CVPRW,
  year={2017}
}

@inproceedings{gu2019div8k,
  title={Div8k: Diverse 8k resolution image dataset},
  author={Gu, Shuhang and Lugmayr, Andreas and Danelljan, Martin and Fritsche, Manuel and Lamour, Julien and Timofte, Radu},
  booktitle=ICCVW,
  year={2019}
}

@inproceedings{timofte2017ntire,
  title={Ntire 2017 challenge on single image super-resolution: Methods and results},
  author={Timofte, Radu and Agustsson, Eirikur and Van Gool, Luc and Yang, Ming-Hsuan and Zhang, Lei},
  booktitle=CVPRW,
  year={2017}
}

@article{karras2019style,
  title={A Style-Based Generator Architecture for Generative Adversarial Networks},
  author={Karras, Tero},
  journal={ArXiv preprint},
  year={2019}
}

@inproceedings{cai2019toward,
  title={Toward real-world single image super-resolution: A new benchmark and a new model},
  author={Cai, Jianrui and Zeng, Hui and Yong, Hongwei and Cao, Zisheng and Zhang, Lei},
  booktitle=ICCV,
  year={2019}
}

@inproceedings{wei2020component,
  title={Component divide-and-conquer for real-world image super-resolution},
  author={Wei, Pengxu and Xie, Ziwei and Lu, Hannan and Zhan, Zongyuan and Ye, Qixiang and Zuo, Wangmeng and Lin, Liang},
  booktitle=ECCV,
  year={2020},
}

@article{wang2004ssim,
  title={Image quality assessment: from error visibility to structural similarity},
  author={Wang, Zhou and Bovik, Alan C and Sheikh, Hamid R and Simoncelli, Eero P},
  journal={TIP},
  year={2004},
}

@inproceedings{ke2021musiq,
  title={Musiq: Multi-scale image quality transformer},
  author={Ke, Junjie and Wang, Qifei and Wang, Yilin and Milanfar, Peyman and Yang, Feng},
  booktitle=ICCV,
  year={2021}
}

@article{sun2023improving,
  title={Improving the stability and efficiency of diffusion models for content consistent super-resolution},
  author={Sun, Lingchen and Wu, Rongyuan and Liang, Jie and Zhang, Zhengqiang and Yong, Hongwei and Zhang, Lei},
  journal={ArXiv preprint},
  year={2023}
}

@inproceedings{yang2022maniqa,
  title={Maniqa: Multi-dimension attention network for no-reference image quality assessment},
  author={Yang, Sidi and Wu, Tianhe and Shi, Shuwei and Lao, Shanshan and Gong, Yuan and Cao, Mingdeng and Wang, Jiahao and Yang, Yujiu},
  booktitle=CVPR,
  year={2022}
}

@inproceedings{wang2023clipiqa,
  title={Exploring clip for assessing the look and feel of images},
  author={Wang, Jianyi and Chan, Kelvin CK and Loy, Chen Change},
  booktitle=AAAI,
  year={2023}
}

@inproceedings{blau2018perception,
  title={The perception-distortion tradeoff},
  author={Blau, Yochai and Michaeli, Tomer},
  booktitle=CVPR,
  year={2018}
}

@inproceedings{jinjin2020pipal,
  title={Pipal: a large-scale image quality assessment dataset for perceptual image restoration},
  author={Jinjin, Gu and Haoming, Cai and Haoyu, Chen and Xiaoxing, Ye and Ren, Jimmy S and Chao, Dong},
  booktitle=ECCV,
  year={2020},
}

@inproceedings{duan2025dit4sr,
  title={Dit4sr: Taming diffusion transformer for real-world image super-resolution},
  author={Duan, Zheng-Peng and Zhang, Jiawei and Jin, Xin and Zhang, Ziheng and Xiong, Zheng and Zou, Dongqing and Ren, Jimmy S and Guo, Chunle and Li, Chongyi},
  booktitle={ICCV},
  year={2025}
}

@inproceedings{wu2024qalign,
  title={Q-Align: Teaching LMMs for Visual Scoring via Discrete Text-Defined Levels},
  author={Wu, Haoning and Zhang, Zicheng and Zhang, Weixia and Chen, Chaofeng and Li, Chunyi and Liao, Liang and Wang, Annan and Zhang, Erli and Sun, Wenxiu and Yan, Qiong and Min, Xiongkuo and Zhai, Guangtai and Lin, Weisi},
  booktitle={ICML},
  year={2024},
}

@article{batifol2025flux,
  title={FLUX. 1 Kontext: Flow Matching for In-Context Image Generation and Editing in Latent Space},
  author={Batifol, Stephen and Blattmann, Andreas and Boesel, Frederic and Consul, Saksham and Diagne, Cyril and Dockhorn, Tim and English, Jack and English, Zion and Esser, Patrick and Kulal, Sumith and others},
  journal={ArXiv preprint},
  year={2025}
}

@article{loshchilov2017decoupled,
  title={Decoupled weight decay regularization},
  author={Loshchilov, Ilya and Hutter, Frank},
  journal={ArXiv preprint},
  year={2017}
}

@article{zhang2025timeawarestepdiffusionnetwork,
    title={Time-Aware One Step Diffusion Network for Real-World Image Super-Resolution}, 
    author={Tainyi Zhang and Zheng-Peng Duan and Peng-Tao Jiang and Bo Li and Ming-Ming Cheng and Chun-Le Guo and Chongyi Li},
    journal={ArXiv preprint},
    year={2025}
}

@inproceedings{hu2022lora,
  title={Lora: Low-rank adaptation of large language models.},
  author={Hu, Edward J and Shen, Yelong and Wallis, Phillip and Allen-Zhu, Zeyuan and Li, Yuanzhi and Wang, Shean and Wang, Lu and Chen, Weizhu and others},
  booktitle={ICLR},
  year={2022}
}

@article{xue2025dancegrpo,
  title={DanceGRPO: Unleashing GRPO on Visual Generation},
  author={Xue, Zeyue and Wu, Jie and Gao, Yu and Kong, Fangyuan and Zhu, Lingting and Chen, Mengzhao and Liu, Zhiheng and Liu, Wei and Guo, Qiushan and Huang, Weilin and others},
  journal={ArXiv preprint},
  year={2025}
}

@inproceedings{zhang2021designing,
  title={Designing a practical degradation model for deep blind image super-resolution},
  author={Zhang, Kai and Liang, Jingyun and Van Gool, Luc and Timofte, Radu},
  booktitle={ICCV},
  year={2021}
}

@inproceedings{rafailov2023direct,
  title={Direct preference optimization: Your language model is secretly a reward model},
  author={Rafailov, Rafael and Sharma, Archit and Mitchell, Eric and Manning, Christopher D and Ermon, Stefano and Finn, Chelsea},
  booktitle={NeurIPS},
  year={2023}
}

@article{zheng2025diffusionnft,
  title={Diffusionnft: Online diffusion reinforcement with forward process},
  author={Zheng, Kaiwen and Chen, Huayu and Ye, Haotian and Wang, Haoxiang and Zhang, Qinsheng and Jiang, Kai and Su, Hang and Ermon, Stefano and Zhu, Jun and Liu, Ming-Yu},
  journal={ArXiv preprint},
  year={2025}
}

@inproceedings{wallace2024diffusion,
  title={Diffusion model alignment using direct preference optimization},
  author={Wallace, Bram and Dang, Meihua and Rafailov, Rafael and Zhou, Linqi and Lou, Aaron and Purushwalkam, Senthil and Ermon, Stefano and Xiong, Caiming and Joty, Shafiq and Naik, Nikhil},
  booktitle={CVPR},
  year={2024}
}

@inproceedings{sun2024pisasr,
  title={Pixel-level and Semantic-level Adjustable Super-resolution: A Dual-LoRA Approach},
  author={Sun, Lingchen and Wu, Rongyuan and Ma, Zhiyuan and Liu, Shuaizheng and Yi, Qiaosi and Zhang, Lei},
  booktitle={CVPR},
  year={2025}
}

@article{wu2025omgsr,
  title={OMGSR: You Only Need One Mid-timestep Guidance for Real-World Image Super-Resolution},
  author={Wu, Zhiqiang and Sun, Zhaomang and Zhou, Tong and Fu, Bingtao and Cong, Ji and Dong, Yitong and Zhang, Huaqi and Tang, Xuan and Chen, Mingsong and Wei, Xian},
  journal={ArXiv preprint},
  year={2025}
}

@article{lin2025hypir,
  title={Harnessing diffusion-yielded score priors for image restoration},
  author={Lin, Xinqi and Yu, Fanghua and Hu, Jinfan and You, Zhiyuan and Shi, Wu and Ren, Jimmy S and Gu, Jinjin and Dong, Chao},
  journal={ArXiv preprint},
  year={2025}
}

@article{wu2025vqr1,
  title={VisualQuality-R1: Reasoning-Induced Image Quality Assessment via Reinforcement Learning to Rank},
  author={Wu, Tianhe and Zou, Jian and Liang, Jie and Zhang, Lei and Ma, Kede},
  journal={ArXiv preprint},
  year={2025}
}

@article{cao2025analytical,
  title={Analytical survey of learning with low-resource data: From analysis to investigation},
  author={Cao, Xiaofeng and Xu, Mingwei and Yu, Xin and Yao, Jiangchao and Ye, Wei and Huang, Shengjun and Zhang, Minling and Tsang, Ivor and Ong, Yew-Soon and Kwok, James T and others},
  journal={ACM Computing Surveys},
  year={2025},
}

@article{wu2025exploring,
  title={Exploring contextual priors for real-world image super-resolution},
  author={Wu, Shixiang and Dong, Chao and Qiao, Yu},
  journal={Computational Visual Media},
  year={2025},
}

@article{wang2024super,
  title={Super-resolution reconstruction of single image for latent features},
  author={Wang, Xin and Yan, Jing-Ke and Cai, Jing-Ye and Deng, Jian-Hua and Qin, Qin and Cheng, Yao},
  journal={Computational Visual Media},
  year={2024},
}

@article{yang2025jvcsr+,
  title={JVCSR+: Adaptively learned video compressive sensing reconstruction with joint in-loop reference enhancement and out-loop super-resolution},
  author={Yang, Jian and Xu, Jiayao and Pham, Chi Do-Kim and Zhou, Jinjia},
  journal={Computational Visual Media},
  year={2025},
}

@article{song2025wdfsr,
  title={WDFSR: Normalizing flow based on the wavelet-domain for super-resolution},
  author={Song, Chao and Li, Shaobang and Li, Frederick WB and Yang, Bailin},
  journal={Computational Visual Media},
  year={2025},
}
}
\end{document}